\documentclass{article}

\usepackage{bbm}
\usepackage{natbib}
\usepackage{verbatim}

\usepackage{subfigure}
\usepackage[preprint]{neurips_2021}

\usepackage[utf8]{inputenc} 
\usepackage[T1]{fontenc}    
\usepackage{hyperref}       
\usepackage{url}            
\usepackage{booktabs}       
\usepackage{amsfonts}       
\usepackage{nicefrac}       
\usepackage{microtype}      
\usepackage{xcolor}         
\usepackage[ruled,vlined]{algorithm2e} 

\usepackage{amsfonts}
\usepackage{algorithmic} 
\usepackage{amsmath}
\usepackage{amsthm}
\usepackage{amssymb}
\usepackage[pdftex]{graphicx}
\usepackage{enumitem}

\newtheorem{theorem}{Theorem}
\newtheorem{lemma}{Lemma}
\newtheorem{definition}{Definition} 
 
\newtheorem{assumption}{Assumption}

\newcommand{\xqedhere}[2]{%
\rlap{\hbox to#1{\hfil\llap{\ensuremath{#2}}}}}

\title{Cascading Bandit under Differential Privacy}

\author{Kun Wang\textsuperscript{1},\quad Jing Dong\textsuperscript{2},\quad Baoxiang Wang\textsuperscript{3} \quad Shuai Li\textsuperscript{1},\quad Shuo Shao\textsuperscript{1}\\
\textsuperscript{1}Shanghai Jiao Tong University, \\
\textsuperscript{2}University of Michigan, \\
\textsuperscript{3}The Chinese University of Hong Kong, Shenzhen.\\
{\tt\small wangkun8512@sjtu.edu.cn},\quad{\tt\small jindong@umich.edu}, \quad{\tt\small bxiangwang@gmail.com}, \\
\quad{\tt\small shuaili8@sjtu.edu.cn},\quad{\tt\small shuoshao@sjtu.edu.cn}
}

\begin{document}

\maketitle

\begin{abstract}
This paper studies \emph{differential privacy (DP)} and \emph{local differential privacy (LDP)} in cascading bandits. Under DP, we propose an algorithm which guarantees $\epsilon$-indistinguishability and a regret of $\mathcal{O}((\frac{\log T}{\epsilon})^{1+\xi})$ for an arbitrarily small $\xi$. This is a significant improvement from the previous work of $\mathcal{O}(\frac{\log^3 T}{\epsilon})$ regret.  Under ($\epsilon$,$\delta$)-LDP, we relax the $K^2$ dependence through the tradeoff between privacy budget $\epsilon$ and error probability $\delta$, and obtain a regret of $\mathcal{O}(\frac{K\log (1/\delta) \log T}{\epsilon^2})$, where $K$ is the size of the arm subset. This result holds for both Gaussian mechanism and Laplace mechanism by analyses on the composition. Our results extend to combinatorial semi-bandit. We show respective lower bounds for DP and LDP cascading bandits. Extensive experiments corroborate our theoretic findings.
\end{abstract}

\section{Introduction}

There exists a rich literature on multi-armed bandits (MAB) as it captures a most fundamental problem in sequential decision making - the exploration-exploitation dilemma \citep{auer2002finite,auer2002nonstochastic,DBLP:journals/ftml/BubeckC12,besbes2014stochastic}. Despite its simpleness, the MAB problem is found use in many important applications such as online advertising and clinical trials. In the original stochastic case of MAB, an agent is presented with $K$ arms and is asked to pull an arm at each round through a finite-time horizon. Depending on the agent's choice, the agent will receive a reward and its goal is set to maximizing the cumulative reward. The natural choice of performance metric is thus the difference between the optimal reward possible and the actual reward received by the agent, which is termed \emph{regret} in bandit literature.

The original stochastic case of the multi-armed bandits model, though powerful, is insufficient to cope with the complexity of real applications. The reward feedback is rarely as straightforward as depicted in stochastic MAB. The cascading bandit problem is then proposed with the intention to better model complicated situations common in applications such as recommendation systems and search engines \citep{kveton2015cascading,NIPS2015_1f50893f,DBLP:conf/uai/ZongNSKWK16,li2016contextual,cheung2019thompson,wang2021conservative}. This variant of MAB gives a realistic depiction of user behavior when searching for attractive things. Cascading bandit is also a sequential learning process, where in each round the agent recommends a list of arms to the user. The user checks from the start of the list and stops at the first attractive item, which is manifested by clicks in web recommendation. Then the agent receives the feedback in the form of user's click information.

While the cascading bandit model greatly assists the development of real applications such as recommendation systems, it raises concern of privacy. Many applications rely heavily on this sensitive user data which reveals the preference of a particular user. If additional measures were not taken, one can easily get this information by observing the difference of the algorithm's output \citep{dwork2006calibrating}. For example, in a shopping website, users' click information and browsing history show the preference for some commodities, and companies can use this information to conduct price discrimination. Since first proposed in 2006, \emph{differential privacy (DP)} becomes a standard in privacy notion because of its well-defined measure of privacy \citep{dwork2006calibrating,dwork2010differential,dwork2014algorithmic,abadi2016deep}. Modifying algorithms to make them possess privacy guarantees is emergent nowadays. An algorithm is regarded to own the capability of protecting differential privacy if the difference between outputs of this algorithm over two adjacent inputs is insignificant, as desired.

\paragraph{Technical difficulty:}

First, the cascading bandits involve a non-linear reward function. Directly injecting noise into the data may lead to much larger regret. Under the canonical definition of differential privacy, bandits algorithms are known to enjoy an upper bound of $\mathcal{O}(\log^3 T)$ \citep{mishra2015nearly, chen2020locally}. Compare to the well known $\mathcal{O}(\log T)$ non private upper bound \citep{auer2002finite,chen2013combinatorial,kveton2015cascading}, an open question remains on whether it is possible to close this gap. 

Moreover, compared to the definition of differential privacy, there is a much stricter definition of privacy guarantee known as \emph{local differential privacy (LDP)}. Under LDP, the data has to be injected with noise before collecting the data and thus it does not need a trusted center, which results in a stronger privacy guarantee. 
There are work on the design of bandit algorithms under LDP \citep{chen2020locally,ren2020multi}. However, in combinatorial (including cascading) bandits, if one directly injects noise to the data and modifies the algorithm in each round as before, the regret scales as $\mathcal{O}(K^2)$, where $K$ is the maximum number of chosen arms in each round. This is a notorious side-effect. In the work of \citet{chen2020locally} for combinatorial bandits, they discard information in each round that does not influence the order of the regret to avoid the dependence on $K^2$. Nevertheless, in real applications, the data is scarce and valuable and need to be well stored to be used in other information analysis tasks. In this case, the natural problem to protect privacy without discarding data and still enjoying a preferable regret bound remains open.


\paragraph{Our contribution:}

In this paper, we overcome the difficulties and solve the problems. Under DP, through analyzing the dominant term in the regret, we utilize the post-processing lemma from \citet{dwork2014algorithmic} and a tighter bound for private continual release from \citet{chan2011private}, proving the $\mathcal{O}(\log^{1+\xi}T)$ regret upper bound, for an arbitrarily small $\xi$. This matches the lower bound up to a $\mathcal{O}(\log^{\xi} T)$ factor. Under LDP, we relax the regret dependence on $K^2$ through the balance between privacy budget $\epsilon$ and error probability $\delta$. This holds for both Laplace mechanism by analyses on composition and Gaussian mechanism. The main contributions of this paper are summarized as follows:

\begin{itemize}[leftmargin=*]
    \item For the cascading bandit problem under $\epsilon$-DP, we propose novel algorithms with regret bound of $\mathcal{O}((\frac{L\log T}{\epsilon})^{1+\xi})$. It matches the lower bound up to an arbitrary small poly-log factor.
    \item For the cascading bandit problem under ($\epsilon,\delta)$-LDP, we propose two algorithms upper bounded by $\mathcal{O}(\frac{K\log 1/\delta \log T}{\epsilon^2})$ which do not discard information. This holds for both Laplace mechanism and Gaussian mechanism.
    \item We extend our mechanism to combinatorial semi-bandit and proved a $\mathcal{O}(\frac{K\log 1/\delta \log T}{\epsilon^2})$ regret, which is of the same order of the regret bound as we achieved in the cascading bandit setting under LDP.
    \item We give two lower bounds, $\Omega(\frac{(L-K)\log T}{\Delta \epsilon^2})$ and $\Omega((\frac{1}{\Delta}+\frac{1}{\epsilon})(L-K)\log T)$, for cascading bandits under LDP and DP, respectively.
\end{itemize}

\section{Related work}
\label{section 2}

\paragraph{Cascading bandit} The cascading bandits problem was first proposed by \citet{kveton2015cascading}, in which a $\mathcal{O}(\log T)$ upper bound by UCB-based methods and a matching lower bound were presented. This problem is then extended to its linear variant  by \citet{DBLP:conf/uai/ZongNSKWK16}, in which a norm bound for contexts from \citet{abbasi2011improved} is employed and a $\mathcal{\tilde{O}}(Kd\sqrt{T})$ upper bound is presented. The combinatorial semi-bandit, which resembles cascading bandit in many aspects is first introduced by \citet{chen2013combinatorial}. They also give corresponding $\mathcal{O}(K\log T)$ upper bound.

\paragraph{Differential privacy} 
\citet{chan2011private} first study the private continual release of statistics, which provides the basis for research for multi-armed bandits. \citet{mishra2015nearly} give an $\mathcal{O}(\frac{K\log^3T}{\epsilon})$ upper bound for the differential private bandit for the first time, where $K$ is the number of arms. \citet{tossou2016algorithms} tighten the upper bound to $\mathcal{O}(\log T)$ but under a relaxed definition of privacy. \citet{tossou2017achieving} also study the adversarial bandit under differential privacy and give a first $\mathcal{O}(\frac{\sqrt{T}\log T}{\epsilon})$ upper regret bound. \citet{NEURIPS2018_a1d7311f} give first lower bound for multi-armed bandit with DP. \citet{basu2019differential} give essential lower bounds for different privacy definitions.

\paragraph{Local differential privacy} Under LDP, \citet{ren2020multi} use the Laplace noise and provide the first $\mathcal{O}(\frac{\log T}{\epsilon^2})$ upper bound for the multi-armed bandit. \citet{chen2020locally} study the combinatorial semi-bandit under LDP and give the first $\mathcal{O}(\frac{\log T}{\epsilon^2})$ regret bound for the combinatorial bandit. They use the trick, each round only update the worse arm in the group, to avoid dependency on $K$, which is the size of each round's chosen arms.

\section{Problem formulation}
\label{section 3}

\subsection{Cascading bandit}
A cascading bandit problem can be denoted by a tuple $B=(E,P,K)$, where $E=\{1,...,L\}$ is a set of $L$ base items, $P$ is a probability distribution over $[0,1]^E$ and its expectation is $(\bar{w})_{i=1}^L$. $K \leq L$ represents the number of recommended items each time. Through a time horizon of $T$, at each round $t$, $w_t \in [0,1]^E$ is instant Bernoulli reward drawn from $P$. It indicates user's preference by $w_t(e)=1$ if and only if item $e$ attracts user at time $t$.

The problem proceeds iteratively, at each round $t$, the algorithm is asked to recommend a list of items $A_t=(a_1^t,a_2^t,...,a_K^t) \in \Pi_K(E)$. When the users receive the recommended list, he reviews the list from the top and stop at first attractive item $C_t$. The feedback to the algorithm is $(w_t)_{i=1}^{C_t}$, where $C_t=\arg \min \{1\leq k \leq K:w_t(a_k^t)=1\}$ with the assumption that $\arg \min \emptyset = \infty$. The reward function $f(A,w)=1-\prod_{k=1}^K (1-w(a_k))$. Note that, $w_t(a_k^t)=\mathbbm{1}\{C_t=k\}\quad k=1,...,\min\{C_t,K\}.$ We say item $e$ is observed at time $t$ if $e=a_k^t$ for some $1\leq k \leq \min\{C_t,K\}$. The cumulative regret $R(T)$ is defined as 
$$\forall T > 1,\quad R(T)=\mathbb{E}[\sum_{t=1}^T R(A_t,w_t)]\,,$$ 
where $R(A_t,w_t)=f(A^\ast,w_t)-f(A_t,w_t)$ is the instant stochastic regret at time $t$ and 
$A^{\ast}={\arg\max}_{A \in \Pi_K(E)} f(A,\bar{w}) \,.$ Next, we maintain two mild assumptions that are common among cascading bandits literature \citep{kveton2015cascading}.
\begin{assumption}
\label{asmp:asmp1}
The weights $w$ are distributed as, $P(w)=\prod_{e \in E}P_e(w(e))\,,$ where $P_e$ is a Bernoulli distribution with mean $\bar{w}(e)\,$.
\end{assumption}
\begin{assumption}
\label{ass2}
The optimal action is unique, i.e. $\Delta_{min}=\min_{e \in [L]\backslash A^\ast}\{\bar{w}^\ast-\bar{w}(e)\}>0\,.$
\end{assumption}

With Assumption \ref{asmp:asmp1}, one can easily verify that $\mathbb{E}[f(A,w)]=f(A,\bar{w})$. This make it possible to decompose the regret to a solvable form. The Assumption \ref{ass2} is necessary for any problem-dependent regrets. Without loss of generality, for all proofs, we assume $w_1 \geq w_2 \geq ... \geq w_L$.

\subsection{(Locally) differential privacy}
We study the problem of cascading bandits under both the classic $(\epsilon,\delta)$-differential privacy definition and $(\epsilon,\delta)$-local differential privacy. We first give the rigorous definition of differential privacy. 
\begin{definition}[Differential Privacy]
Let $D = \langle x_1, x_2, \dots, x_T \rangle$ be a sequence of data with domain $X^T$. 
Let $A(D) = Y$, where $Y = \langle y_1, y_2, \dots, y_T \rangle \in Y^T$ be $T$ outputs of the randomized algorithm $A$ on input $D$. 
$A$ is said to preserve $(\epsilon,\delta)$-differential privacy, if for any two data sequences $D, D'$ that differ in at most one entry, and for any subset $U \subset Y^T$, it holds that
\[
  P(A(D)\in U) \leq e^\epsilon \cdot P(A(D')\in U)+\delta\,.
\]
If $\delta=0$, then we say the algorithm satisfies $\epsilon$-differential privacy.

\end{definition}
The local differential privacy model requires masking data with noise before the accumulation of data in order to circumvent the need of a trusted center, which leads to a more promising privacy guarantee. Formally, the  $(\epsilon,\delta)$-local differential privacy is defined as follows.
\begin{definition}[Local Differential Privacy]
    A mechanism $A: X \rightarrow Y$ is said to be $(\epsilon,\delta)$-local differential private or $(\epsilon,\delta)$-LDP, if for any $x, x' \in X$, and any (measurable) subset $U \subset Y$, there is 
    \begin{align*}
        P(A(x) \in U) \leqslant e^{\epsilon} \cdot P(A(x') \in U)+\delta\,.
    \end{align*}
If $\delta=0$, then we say the algorithm satisfies $\epsilon$-local differential privacy.
\end{definition}



\section{Cascading bandit under differential privacy}
\label{section 4}

In this section, we study cascading bandit under DP. In this situation, each round, the user's click information is submitted to a database as a trusted center. When the algorithm needs data each round, it receives the noisy data from the database.

The difficulty of this problem is governed by the control of the overall injected noise in the stream data. Allocating the injected noise and adjusting the confidence bound in the algorithm to guarantee better regret is the core in this problem. In the previous work, people use a tree-based mechanism or hybrid mechanism from \citet{chan2011private} to maintain the privacy of the stream data in the MAB setting. It means constructing a tree based on previous data to have a logarithmic number of sum when adding noise. This can control all noises to an acceptable content. However, people either not use best confidence bound or improperly construct the tree over stream, leading to some sub-optimal regret bounds.

In this paper, through careful consideration, we find the order of the regret in the bandit algorithm is dominated by the sequence length of the constructed segment tree. If we see each round's data as a $K$ length vector such as \citet{chen2020locally}, then the sequence length is related to time $t$. In this case, it is unavoidable with the $\log^{2.5}t$ by direct derivation from the utility bound of the hybrid mechanism. It is hard to accept. Nonetheless, if the sequence length is relevant with the number of each arm $T_i$ rather than time $t$, we prove that the overall regret can be bounded by $\log^{1+\xi}t$ regret by a common inequality. In this work, we utilize the post-processing lemma well-known in differential privacy, converting the privacy 
guarantee of algorithm's output to $L$ arms. Then just equally distribute the privacy budget to all $L$ arms. We can control the pulled numbers of any sub-optimal arm to $\log^{1+\xi}t$, which is the dominant term in regret.

We now describe our algorithm under DP. It is inspired by \citet{tossou2016algorithms}, but under a much stronger definition of privacy. First, the agent receives data from hybrid mechanism and calculate the upper confidence bound for each arm. Compared to classic UCB algorithm, we add an extra additional term from the noise from the injected noise. The algorithm then outputs $K$ arms. The agent of the algorithm sends chosen $K$ arms to users. Finally, the users pull arms receiving rewards and insert the rewards to the hybrid mechanism for each arm, respectively. Our method is depicted in Algorithm \ref{alg1}.
\begin{algorithm}[H]
\label{alg1}
	\caption{Cascading-UCB under DP.}
	\KwIn{$\epsilon$, the differential privacy parameter}
	Instantiate $L$ Hybrid Mechanisms with $\epsilon'=\frac{\epsilon}{L}$.\\
	Observe $w_0$\\
	$\forall e \in E:T_0(e)\leftarrow 1,\hat{w}_1(e)\leftarrow w_0(e).$\\
	\For{$t =1,2,...,n$}{

		$\forall e \in E:U_t(e)=\hat{w}+\sqrt{\frac{1.5\log t}{T_{t-1}(e)}}+\frac{3c_1L\log^{1.5}n\log t}{\epsilon T_{t-1}(e)}$\\
		
		Let $a_1^t,...,a_K^t$ be $K$ items with largest private UCBs.\\
		$A_t\leftarrow (a_1^t,...,a_K^t)$\\
		Observe click $C_t \in \{1,...,K,\infty\}$
		
		\For{$k=1,...,\min\{C_t,K\}$}{
			e $\leftarrow a_k^t$\\
			$T_t(e)\leftarrow T_{t-1}(e)+1$\\
			Insert $w_t(e)$ to the hybrid mechanism for arm e.\\
		} 
	}
\end{algorithm}

Next, we introduce two important lemmas to support our algorithm and proof.

\begin{lemma}[\citet{chan2011private}, Utility Bound For Hybrid Mechanism]
\label{lem1}
In the continual release procedure, the hybrid mechanism preserves $\epsilon$-differential privacy and with probability $1-\gamma$, the added noise to the data at time $n$ satisfies following inequality:
$$|Noise| \leq \frac{c_1\log^{1.5}n\log 1/\gamma}{\epsilon}\,,$$
where $c_1$ is a constant.
\end{lemma}

\begin{lemma}[Post-Processing Lemma]
\label{lem2}
If the sequence  $(w_t(e))_{t=1}^{T_e}$ for all arm $e$ is $\epsilon$-differentially private, then the Algorithm 1 is $\epsilon$-differentially private.
\end{lemma}

Based on Lemma \ref{lem1}, we can construct high probability events that estimation outside of this confidence interval happens with arbitrary small probability. This is the basis of UCB algorithms. The Lemma \ref{lem2} ensures Algorithm \ref{alg1} is $\epsilon$-differentially private.


\begin{theorem}
Algorithm 1 guarantees $\epsilon$-DP.
\end{theorem}

\begin{theorem}
The regret of Algorithm 1 is upper bounded by:
$$R(T)\leq \sum_{e=K+1}^L \frac{192}{\Delta_{e,K}} \left(\frac{c_1 L}{\epsilon}\log T\right)^{1+\xi}+\frac{2\pi^2}{3}L+c_2=\mathcal{O}\left(\sum_{e=K+1}^L \left[\frac{L\log T}{\epsilon}\right]^{1+\xi}\right)\,,$$
where $\xi$ is an arbitrary small positive real value and $c_1,c_2$ are constants independent of the problem.
\end{theorem}

To the best of our knowledge, this is the first $\mathcal{O}(\log^{1+\xi} T)$ regret in the common definition of differential privacy. We greatly improve the existing regret bound by the post-processing lemma and the utility bound on hybrid mechanism. Judging from the regret, we get this better dependence on $t$ at the cost of extra dependence on $L$. However, in view of time $t$'s dominance in the bandit setting, we believe this is a great improvement compared to before. Besides, our proof method has the universality. It can be used in improving bound on other bandits model under differential privacy such as combinatorial semi-bandit and basic MAB. We do not give them in this paper. 

\section{Cascading bandit under local differential privacy}
\label{section 5}

Under local differential privacy, each round when the user browses through the list, he directly sends noisy data to the database rather than sending original data and let database inject noise. This circumvents the need of trusted center, so it has much stronger privacy guarantee. In this situation, we need to protect every output at time $t$ and let it satisfy $(\epsilon,\delta)$-local differential privacy. In previous work, \citet{chen2020locally} discard much information each round to reduce the privacy budget in the combinatorial semi-bandit. This is not practical as the importance of information in the real world. Therefore, in this paper, we study if there 
exists a method can both satisfy $(\epsilon,\delta)$-LDP and appreciable regret not discarding information.
 
\subsection{Warm up: Laplace mechanism}

First, we use Laplace mechanism to provide LDP guarantee. Our algorithm is based on previous non-private cascading bandit. In order to guarantee privacy constraint, each round we have to inject noise to the data, so original confidence interval is no longer applicable. Naturally, the confidence set has to expand to adapt to the introduction to new Laplace noises. Then, we give our method in Algorithm \ref{alg2}.

\begin{theorem}
Algorithm 2 guarantees $\epsilon$-LDP.
\end{theorem}

\begin{theorem}
\label{theorem 4}
The regret of Algorithm 2 in time horizon T is upper bounded by:
$$R(T)\leq \sum_{e=K+1}^L \frac{4(\sqrt{1.5}+K/\epsilon \sqrt{24})^2}{\Delta_{e,K} }\log T+\frac{2\pi^2}{3}L =\mathcal{O}\left(\sum_{e=K+1}^L \frac{K^2}{\epsilon^2\Delta_{e,K}} \log T\right)\,.$$
\end{theorem}

Next, we discuss the proof outline of Theorem \ref{theorem 4}: First, the expected mean outside the confidence set is bounded by a small probability because of the sub-Gaussian property of two noises. Then we just consider event in high-probability set. The upper confidence bound has a $\frac{K}{\epsilon}$ factor because of Laplace noise. Then follow the same procedure as non-private cascading bandit \citep{kveton2015cascading}, we get $T_{i} \leq \mathcal{O}(\frac{K^2\log T}{\epsilon^2})$ for all arm $i$. Combing with the decomposed form of regret on account of nonlinear reward function, the final regret is the weighted sum of all $T_i$. Thus the final regret has $\mathcal{O}(\frac{K^2}{\epsilon^2})$ dependence.
	\begin{algorithm}[H]
	\label{alg2}
		\caption{Cascading-UCB under LDP (Laplace Mechanism).}
		\KwIn{$\epsilon$: the differential privacy parameter,$K$: the maximum arm each round recommends. }
		Initialization:\\
		Observe $w_0$\\
		$\forall e \in E:T_0(e)\leftarrow 1,\hat{w}_1(e)\leftarrow w_0(e)$\\
		\For{$t =1,2,...,n$}{
		    $\forall e \in E:U_t(e)=\hat{w}(e)+\sqrt{\frac{1.5\log t}{T_{t-1}(e)}}+\frac{K}{\epsilon}\sqrt{\frac{24 \log t}{T_{t-1}(e)}}$\\
		    
		    Let $a_1^t,...,a_K^t$ be $K$ items with largest private UCBs.\\
		    $A_t\leftarrow (a_1^t,...,a_K^t)$\\
		    Observe click $C_t \in \{1,...,K,\infty\}$
		    
		    \For{$k=1,...,\min\{C_t,K\}$}{
		         e $\leftarrow a_k^t$\\
		         $T_t(e)\leftarrow T_{t-1}(e)+1$\\
		         $\hat{w}_{T_e(e)}(e) \leftarrow \frac{T_{t-1}(e)\hat{w}_{T_{t-1}(e)}(e)+\mathbbm{1}\{C_t=k\}+Lap(\frac{K}{\epsilon}) }{T_t(e)}$
		    
		    } 
		}
 	\end{algorithm}

In our first given Laplace mechanism, the dependence on $K^2$ is a notorious side-effect during the learning process. In recommender system, $K$ means the recommended items to user and it is usually very large. In scenarios where recommendation items come from a stream of information such as news feed, $K$ often tends to infinity. This leads to much larger regret. Therefore, a tough work about reducing dependence on $K$ is in emergency. In subsection (\ref{sub1})(\ref{sub2})(\ref{sub3}), we make great efforts in reducing this dependence on $K$.

\subsection{Improvement: Gaussian mechanism}
\label{sub1}

The Laplace mechanism, though offering promising privacy protection, is quadratically dependent on $K$. This thus makes the Laplace mechanism impractical when $K$ is large. Previous solution to this catastrophic effect is to discard information with negligible effect on the order of regret bound \citep{chen2020locally}. Despite the alleviation of the effect, this method is inaccessible in many situations where data disposal is impossible. We now present an improved algorithm to circumvent this dependency $K$ subtly, which is based on the Gaussian mechanism.

	  	\begin{algorithm}[H]
	  	\label{alg3}
		\caption{Cascading-UCB under LDP(Gaussian Mechanism).}
		\KwIn{$\epsilon$: the differential privacy parameter,$K$: the maximum arm each round recommends. }
		Initialization:$\sigma=\frac{1}{\epsilon}\sqrt{2K \ln \frac{1.25}{\delta}}$\\
		Observe $w_0$.\\
		$\forall e \in E:T_0(e)\leftarrow 1,\hat{w}_1(e)\leftarrow w_0(e).$\\
		\For{$t =1,2,...,n$}{
		    $\forall e \in E:U_t(e)=\hat{w}+\sqrt{\frac{1.5\log t}{T_{t-1}(e)}}+\sigma\sqrt{\frac{2\log (2t^3)}{T_{t-1}(e)}}$\\

		    Let $a_1^t,...,a_K^t$ be $K$ items with largest private UCBs.\\
		    $A_t\leftarrow (a_1^t,...,a_K^t)$\\
		    Observe click $C_t \in \{1,...,K,\infty\}$
		    
		    \For{$k=1,...,\min\{C_t,K\}$}{
		         e $\leftarrow a_k^t$\\
		         $T_t(e)\leftarrow T_{t-1}(e)+1$\\
		         $\hat{w}_{T_e(e)}(e) \leftarrow \frac{T_{t-1}(e)\hat{w}_{T_{t-1}(e)}(e)+\mathbbm{1}\{C_t=k\}+\mathcal{N}(0,\sigma^2) }{T_t(e)}$
		    
		    } 
		}
	\end{algorithm}

The Gaussian mechanism, compared to Laplace mechanism, is sensitive to $L_2$-norm instead of $L_1$-norm. This key property leads to our design of a $\sqrt{K}$-dependent of upper confidence bound. The adjustment to the confidence interval, directly lessen the dominating effect of the interval on the regret bound and eliminate the $K^2$ dependency. We now give the detailed algorithm description for Gaussian mechanism based cascading UCB algorithm with LDP guarantee. Before going on our work, we first introduce a lemma.


\begin{lemma}[\citet{zhao2019reviewing}]
\label{lem3}
Let $\Delta_f=\max \limits_{D,D'} \Vert f(D)-f(D')\Vert_{L_2} $, then $\forall \delta \in (0,1)$, and $\sigma > \frac{\Delta_f}{\epsilon}\sqrt{2\log \frac{1.25}{\delta}}$,$M(D)=f(D)+\mathcal{N}(0,\sigma^2)$ satisfied $(\epsilon,\delta)$-differential privacy.
\end{lemma}

Lemma \ref{lem3} gives the Gaussian mechanism privacy guarantee, if we inject $\mathcal{N}(0,\sigma^2)$ to reward of each item according to lemma. Based on the concentration bound for Gaussian distribution, we get with probability of at least $1-\gamma$,
$u \in [\bar{X}-\sigma \sqrt{\frac{2 \log 2/\gamma}{s}},\bar{X}+\sigma \sqrt{\frac{2 \log 2/\gamma}{s}}]$. By the concentration bound for Gaussian noise, we can use the Optimism in the Face of Uncertainty principle to construct UCB-based method. The bias comes from two noises: the noise from the Bernoulli distribution, this part we can control by sub-Gaussian quality; the noise from the Gaussian noise due to the privacy guarantee, this part can be controlled by the concentration bound for Gaussian distribution. Then follow the same procedure as Laplace mechanism. Our method is depicted in Algorithm \ref{alg3}.


\begin{theorem}
Algorithm 3 guarantees $(\epsilon,\delta)$-LDP.
\end{theorem}

\begin{theorem}
\label{theorem6}
The regret of Algorithm 3 is upper bounded by:
$$R(T)\leq \sum_{e=K+1}^L \frac{2 (2\sqrt{1.5}+\frac{8}{\epsilon}\sqrt{K\log \frac{1.25}{\delta}})^2}{\Delta_{e,K}}\log T=\mathcal{O}\left(\sum_{e=K+1}^L \frac{K}{\epsilon^2\Delta_{e,K}}\log T\right).$$
\end{theorem}

During the proof, the only position where the Gaussian mechanism is different from the Laplace mechanism, is the confidence interval. We update the interval to $\mathcal{O}(\frac{\sqrt{K}}{\epsilon})$ dependence based on concentration bound for Gaussian distribution. As we have discussed in the above, the regret has the linear dependence on $K$. It is a great improvement from the Laplace mechanism. However, more fundamental reasons need to be explored to guide our algorithm's design.

\subsection{Generalization: composition theorem}
\label{sub2}

After careful investigation the regret of Gaussian mechanism, we identify a trade-off between privacy budget $\epsilon$ and error probability $\delta$ in regret. Comparing the regret of Gaussian mechanism and that of Laplace mechanism, the dependence on privacy budget $\epsilon (\frac{\epsilon}{K})$ is exchanged by an additional multiplicative $\delta (\log\frac{1}{\delta})$ regret. This motivation of exchanging for $\epsilon$ by a little $\delta$ is just why advanced composition theorem in differential privacy is proposed. Thus we can use this composition theorem at our situation to reduce the dependence on $K$.

\begin{lemma}[\citet{kairouz2015composition}, Corollary 4.1]
\label{theorem 9}
For any $\epsilon \in (0,0.9]$ and $\delta \in (0,1]$, if the database access mechanism satisfies $(\sqrt{\epsilon^2/4k\log(e+\epsilon/\delta)},\delta/2k)$-differential privacy on each query output, then it satisfies $(\epsilon,\delta)$-differential privacy under k-fold composition.
\end{lemma}

Before, we have to ensure every item $\frac{\epsilon}{K}$-indistinguishable. Now, we just to ensure each item $\frac{\epsilon}{\sqrt{K}}$-indistinguishable based on Lemma \ref{theorem 9}. Thus any $\epsilon$-dp mechanism can achieve half dependence on $K$ at the cost of a $\log 1/\delta$ term in regret while ensuring their $(\epsilon,\delta)$-local differential privacy. We use this composition theorem in our situation. We give an illustrating example with the Laplace mechanism to highlights the impact of the above theorem. By this theorem, we can improve the regret of Laplace mechanism to $\mathcal{O}(K)$ dependence in regret, which is the same as Gaussian mechanism.



\begin{theorem}
Cascading-UCB algorithm using Laplace mechanism with parameter $\epsilon'=\frac{\epsilon}{\sqrt{4K\log(e+\epsilon/\delta)}}$ under K-fold composition each round achieves $\mathcal{O}(\frac{K\log 1/\delta \log T}{\epsilon^2})$ regret while ensuring $(\epsilon,\delta)$-local differential privacy.
\end{theorem}

By the definition of DP, Laplace mechanism that attains $(\epsilon,0)$-DP is also capable of offering $(\epsilon,\delta)$-DP protection. Using the generalized theorem, every item observed in the list masked with a $Lap(\frac{\sqrt{4K\log(e+\epsilon/\delta)}}{\epsilon})$ noise is enough to guarantee $(\epsilon,\delta)$-dp. An UCB algorithm with confidence interval of $\frac{4}{\epsilon}\sqrt{\frac{6K \log (e+\epsilon/\delta)\log t}{T_{t-1}(e)}}+\sqrt{\frac{3\log t}{2T_{t-1}(e)}}$ then suffers only $\mathcal{O}(\frac{K\log 1/\delta \log T}{\epsilon^2})$ regret while ensuring privacy guarantee.

\subsection{Relationship with combinatorial semi-bandit}
\label{sub3}
Our proposed algorithms also shed light on privacy persevering bandits algorithm under combinatorial semi-bandit, which holds great similarity with cascading bandit set up. We take the Gaussian mechanism as an example and give the following theorem on the regret upper bound of UCB algorithm under LDP with combinatorial bandit, though we defer the detailed algorithm description to appendix as Algorithm \ref{algorithm 4}.


\begin{theorem}
Algorithm 4 guarantees $(\epsilon,\delta)$-LDP.
\end{theorem}
\begin{theorem}
The regret of Algorithm 4 is upper bounded by:
$$R(T)\leq \left[\frac{128K\log \frac{1.25}{\delta} \log T}{\min\{\epsilon^2,2\}(f^{-1}(\Delta_{min}))^2}+\frac{\pi^2}{3}+1\right]m \Delta_{max}\,,$$
where is $f$ is strictly increasing function satisfying $B_\infty$ bounded smoothness and $K$ is maximum number of each round chosen arm.
\end{theorem}

\section{Lower bounds}
\label{section 6}

We now discuss the lower bound for cascading bandit under DP and LDP and the optimality of our upper bounds. We first decompose the lower bound for cascading bandit into a weighted sum of any sub-optimal arm's pulled number. Under LDP, this can be further converted into the KL-divergence of two arms and thus implies a $\Omega(1/\epsilon^2)$ dependence according to the results from \citep{basu2019differential}. Under DP, we decompose the number of any sub-optimal arm into two parts: the number of pulls with no privacy guarantee and the number of pulls with the cost of privacy. The following theorems formalize our results. 
\begin{theorem}
For any cascading bandit of $\epsilon$-local differential privacy, the regret of any consistent algorithm $R(T)$ is lower bounded by 
\begin{align*}
  \liminf_{T \to \infty} \frac{R(T)}{\log T} \geq
  \frac{(L - K) p (1 - p)^{K }}{2\min\{4,e^{2\epsilon}\}(e^{\epsilon}-1)^2\Delta}\,.
\end{align*}
Note that when $p=1/K$ and $\epsilon \rightarrow 0$, $\liminf_{T \to \infty} R(T) \geq
\Omega(  \frac{(L - K) }{\Delta \epsilon^2}\log T).$
\end{theorem}

\begin{theorem}
For any cascading bandit of $\epsilon$-differential privacy, the regret of any consistent algorithm $R(T)$ is lower bounded by  
\[
R(T)\geq \Omega\left(\frac{(L-K)(1-p)^{K-1}\log T}{\Delta}+\frac{(L-K)(1-p)^{K-1}\log T}{\epsilon}\right) \,.
\]
\end{theorem}
It can be observed that the definition of privacy protection, whether it is differential privacy or local differential privacy, has negligible effect on the lower bound of cascading bandits. Our proposed algorithm for cascading bandits achieves matching upper bound under the LDP while there exists a $\mathcal{O}(\log^{\xi} T)$ gap under DP.

\section{Empirical results}
\label{section 7}
In this section, we provide extensive empirical results that corroborate our theoretical findings. We conduct experiments under DP and LDP for Gaussian and Laplace mechanism with varying value of $\epsilon$. In order to highlight our improvement, Laplace mechanism uses the form of Algorithm \ref{alg2}. The experiments were proceeded with $L=20, K=4,\delta=10^{-3}$ unless other wise indicated. The time horizon is set to $100000$ for all experiments. To sure reproducible results, each experiment is repeated 10 times and averaged result is presented. Without loss of generality, all noises are multiplied by 0.01 to ensure most of the samples distributed among $[0,1]$.

\begin{figure}[htbp] \label{fig}
\centering
\subfigure[LDP,$\epsilon=0.2$]{ \label{ldp1}
\includegraphics[width=0.31\textwidth]{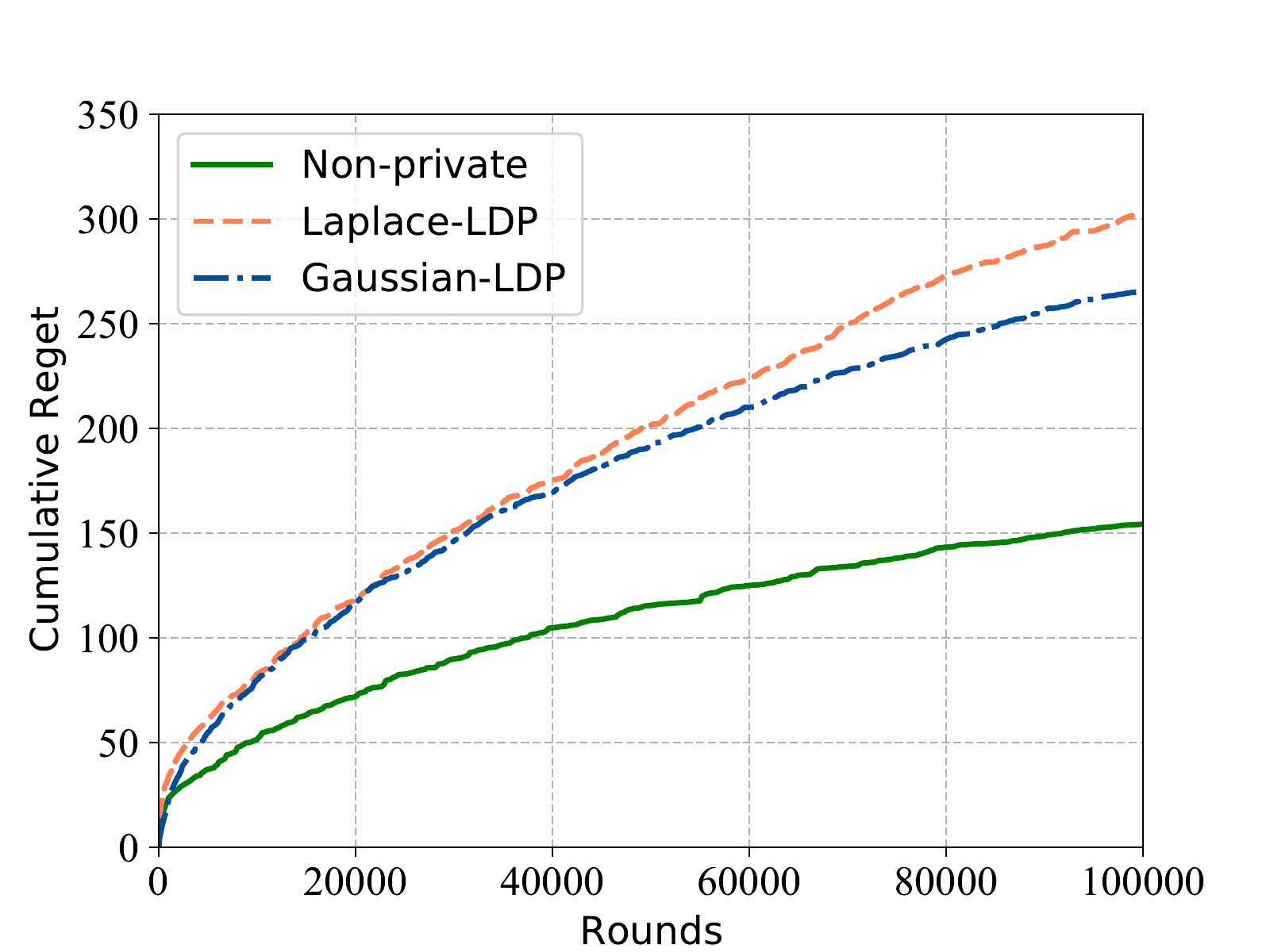}
}
\subfigure[LDP, $\epsilon=0.5$]{  \label{ldp2}
\includegraphics[width=0.31\textwidth]{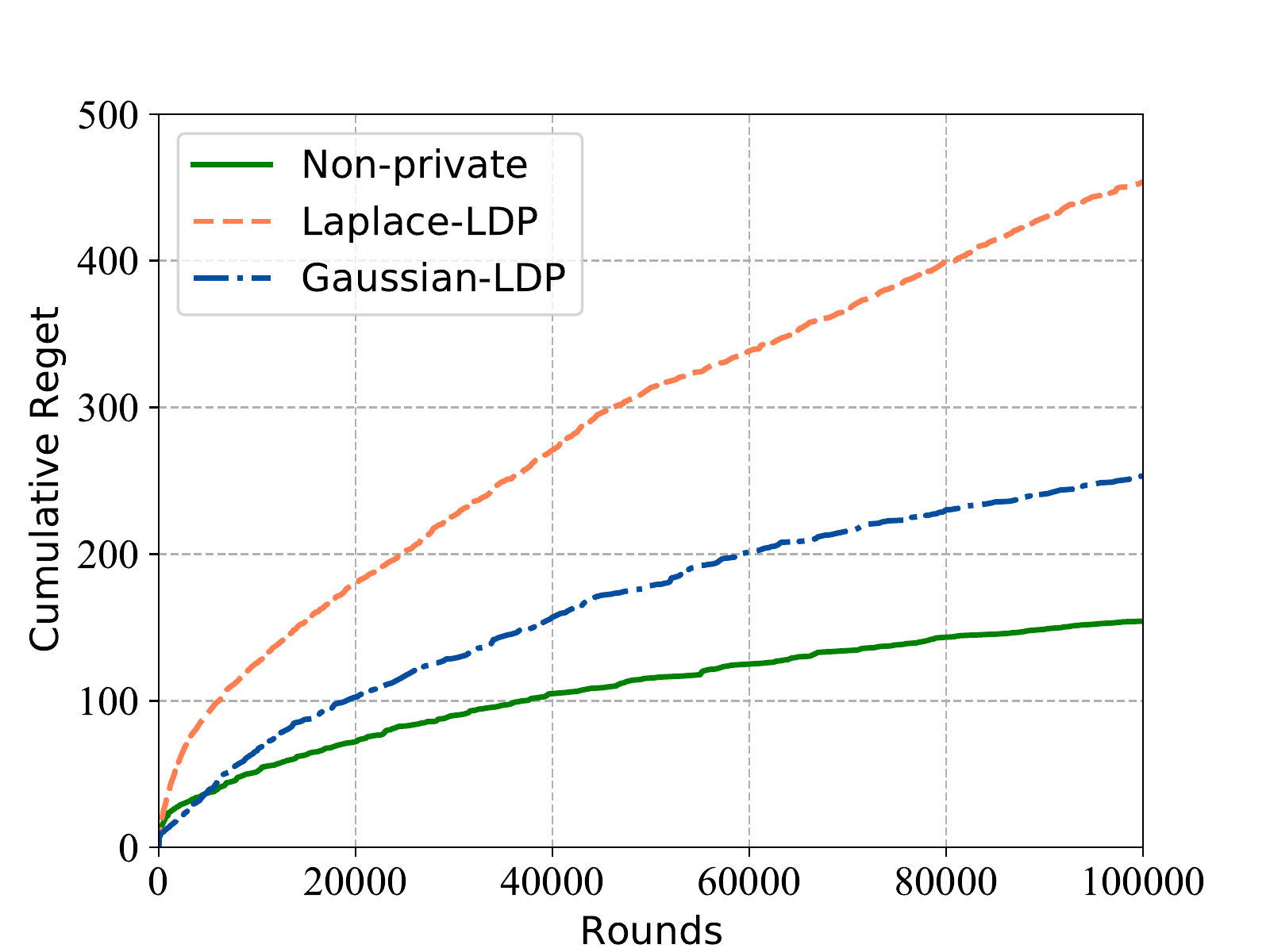}
}
\subfigure[LDP, $\epsilon=1$]{ \label{ldp3}
\includegraphics[width=0.31\textwidth]{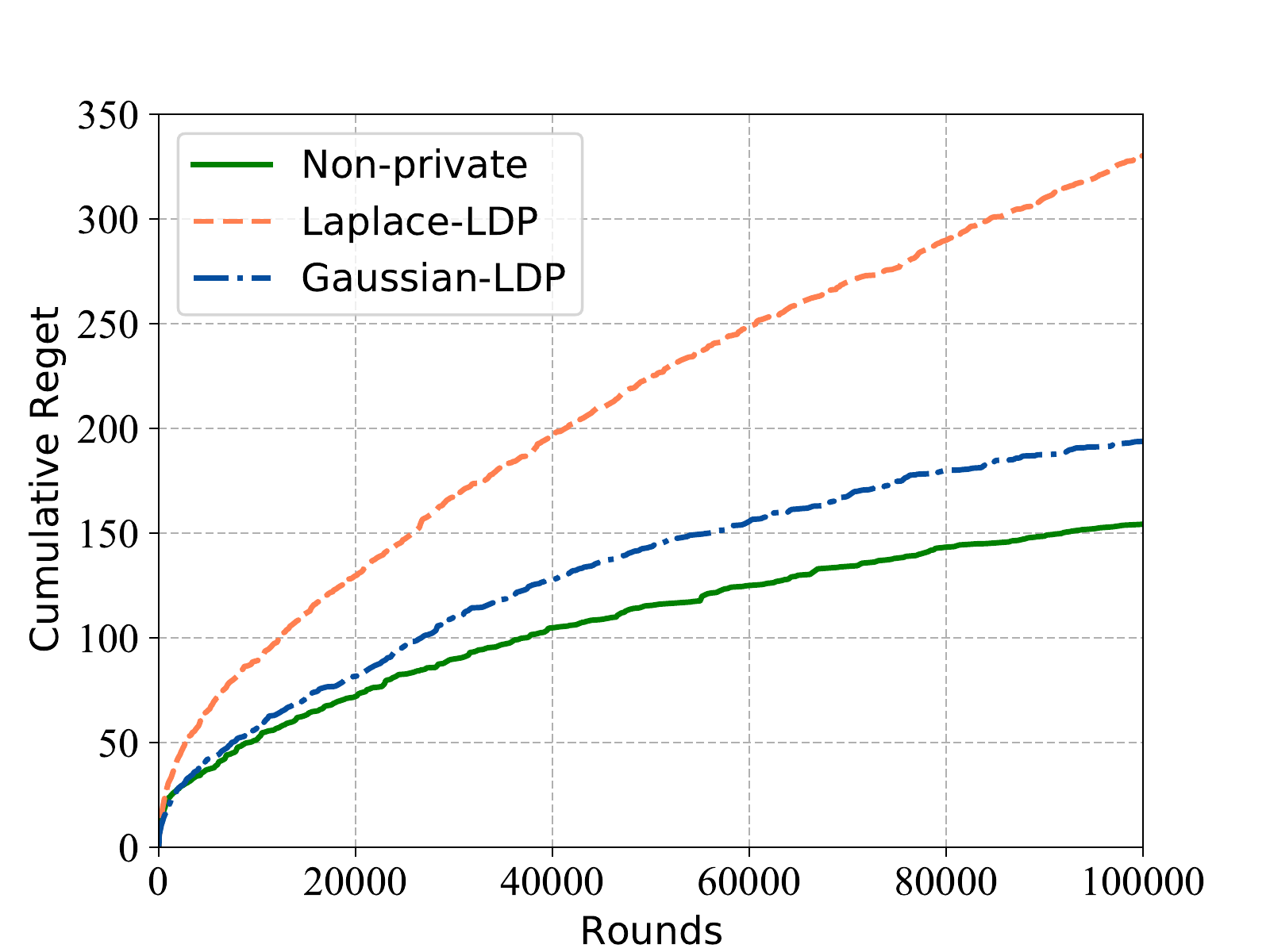}
}

\subfigure[LDP, $\epsilon=2$]{ \label{ldp4}
\includegraphics[width=0.31\textwidth]{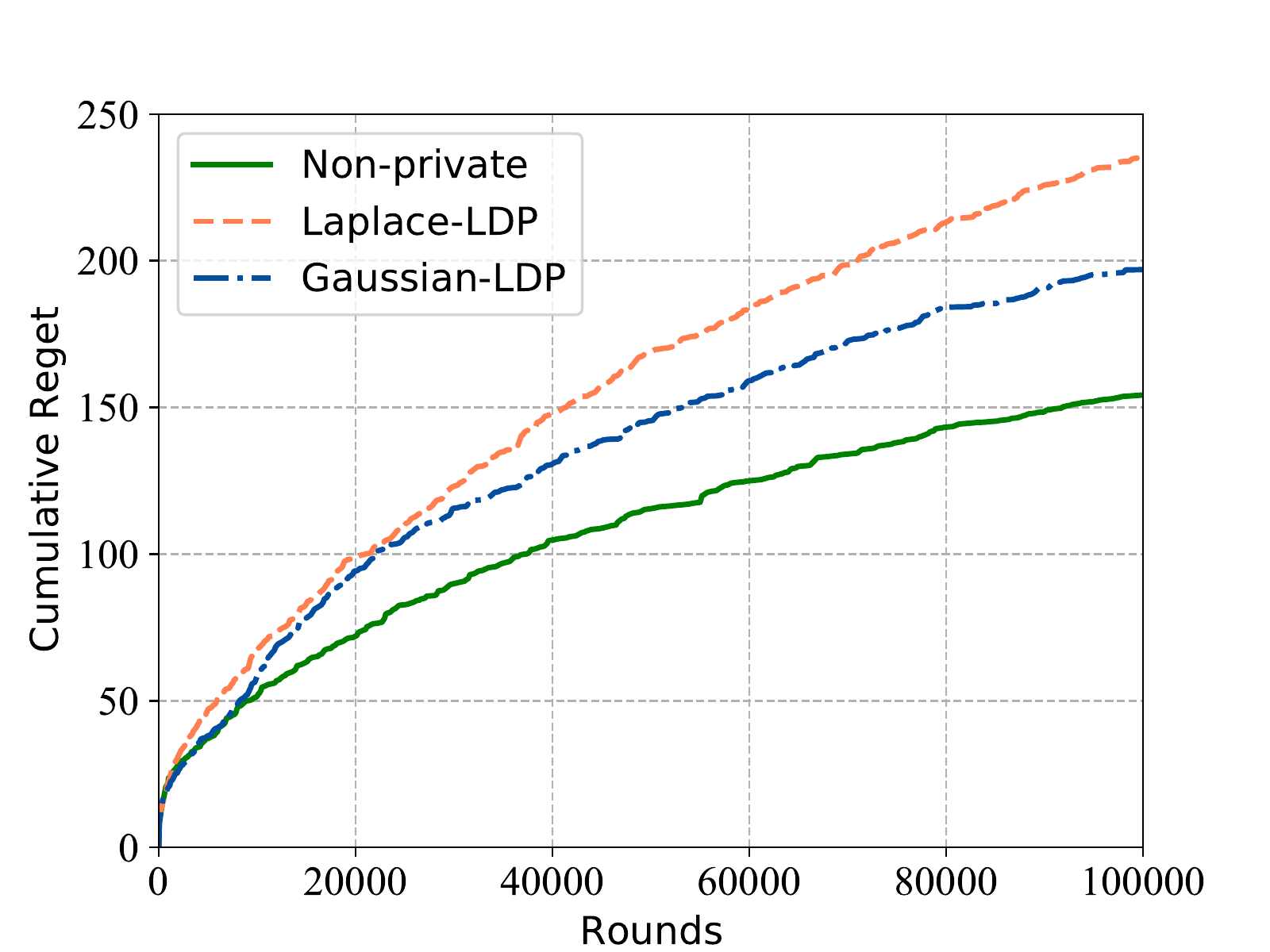}
}
\subfigure[LDP, vary $\epsilon$]{ \label{ldp5}
\includegraphics[width=0.31\textwidth]{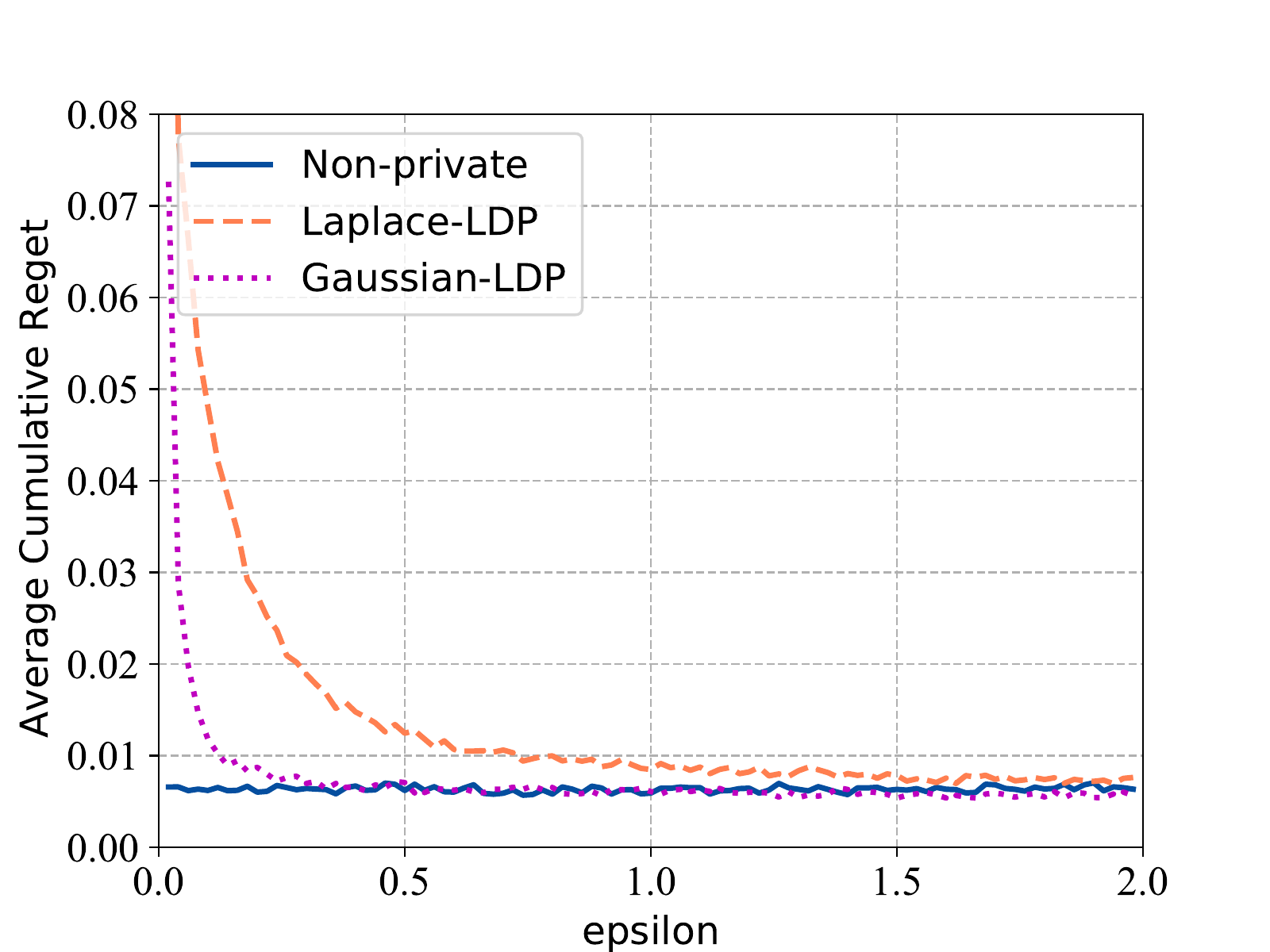}
}
\subfigure[DP, vary $\epsilon$]{ \label{dp1}
\includegraphics[width=0.31\textwidth]{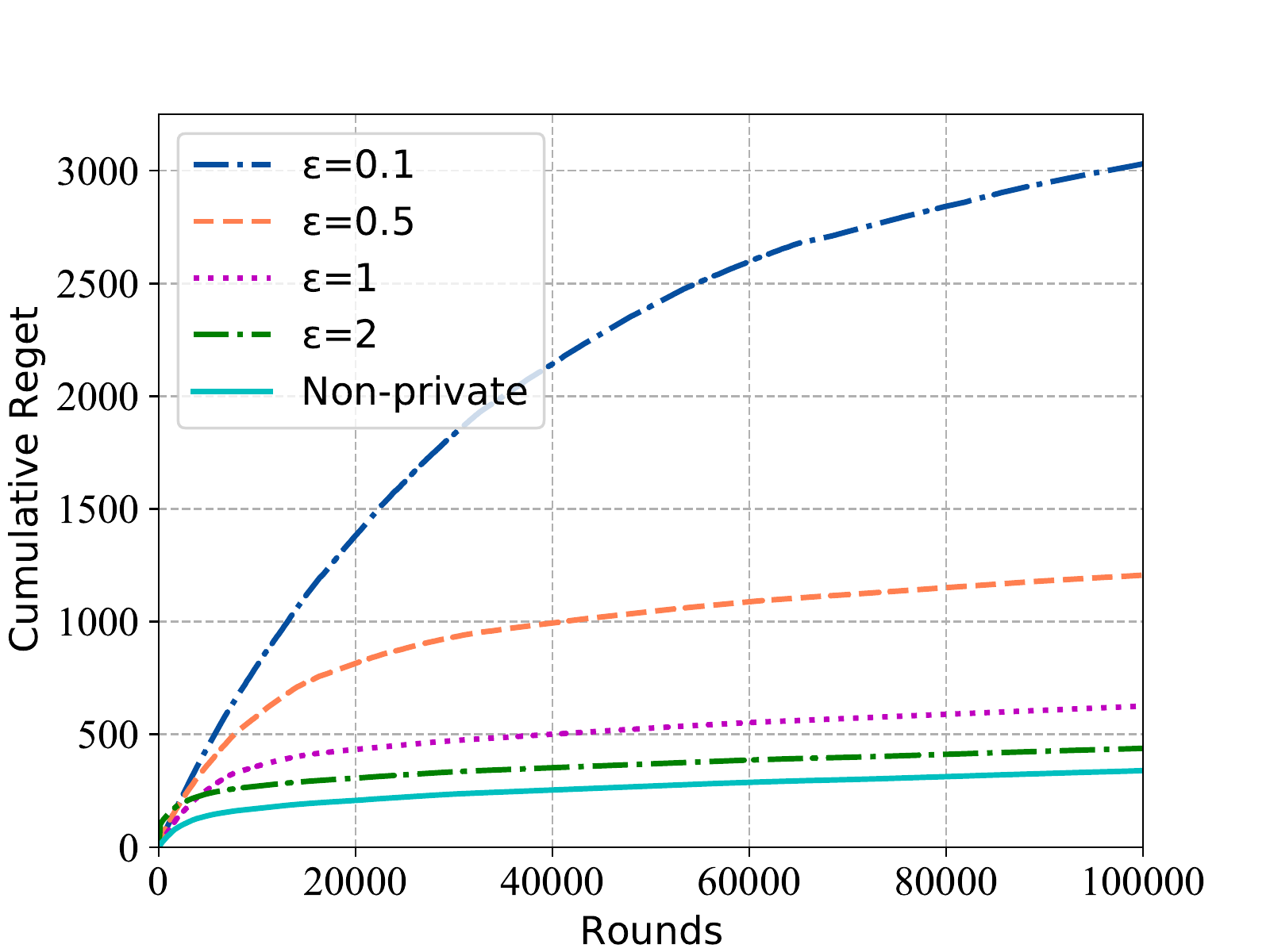}
}
\subfigure[DP, vary $L$]{ \label{dp2}
\includegraphics[width=0.31\textwidth]{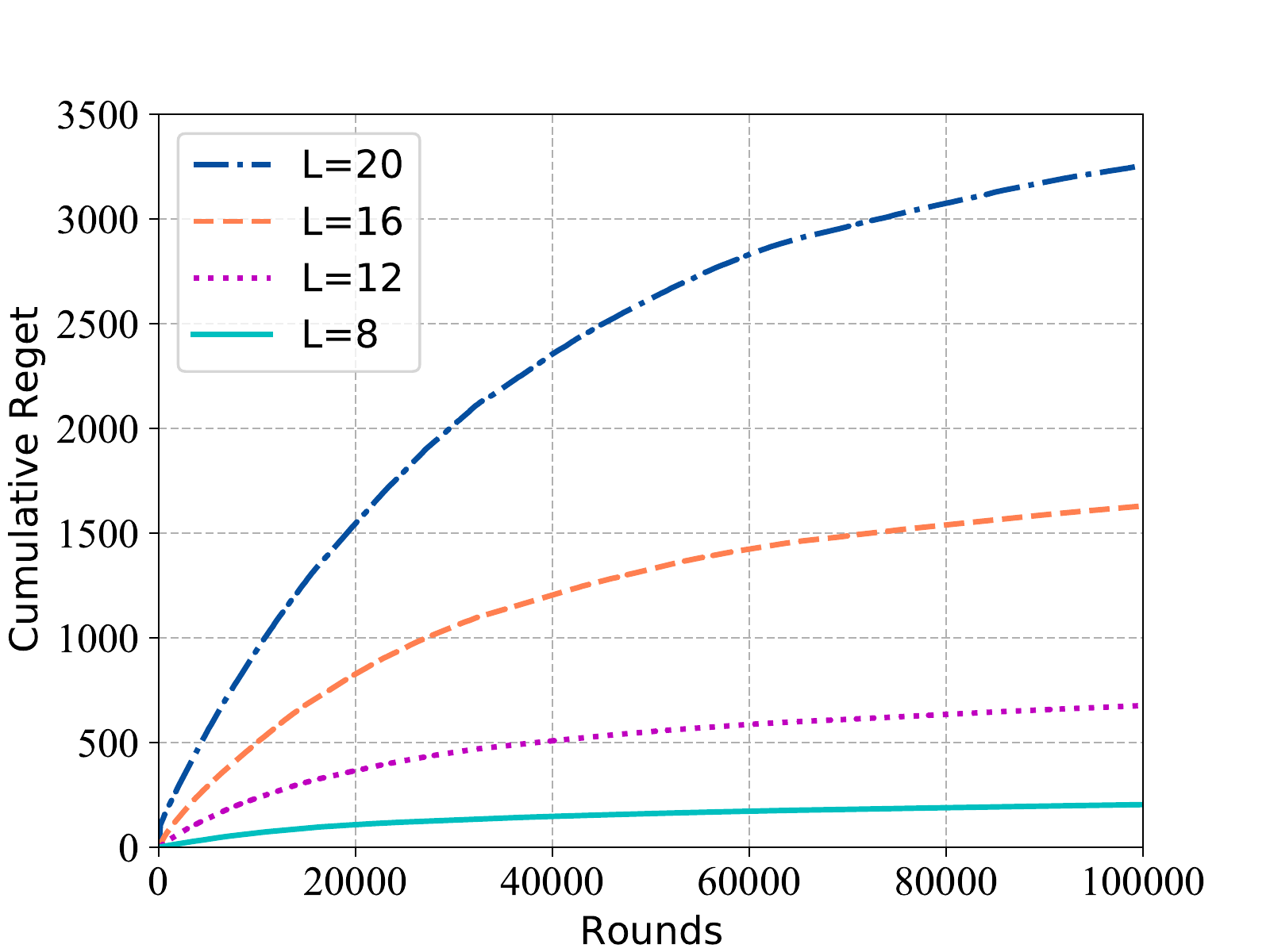}
}
\subfigure[CUCB under LDP, $\epsilon=0.2$]{ \label{cucb1}
\includegraphics[width=0.31\textwidth]{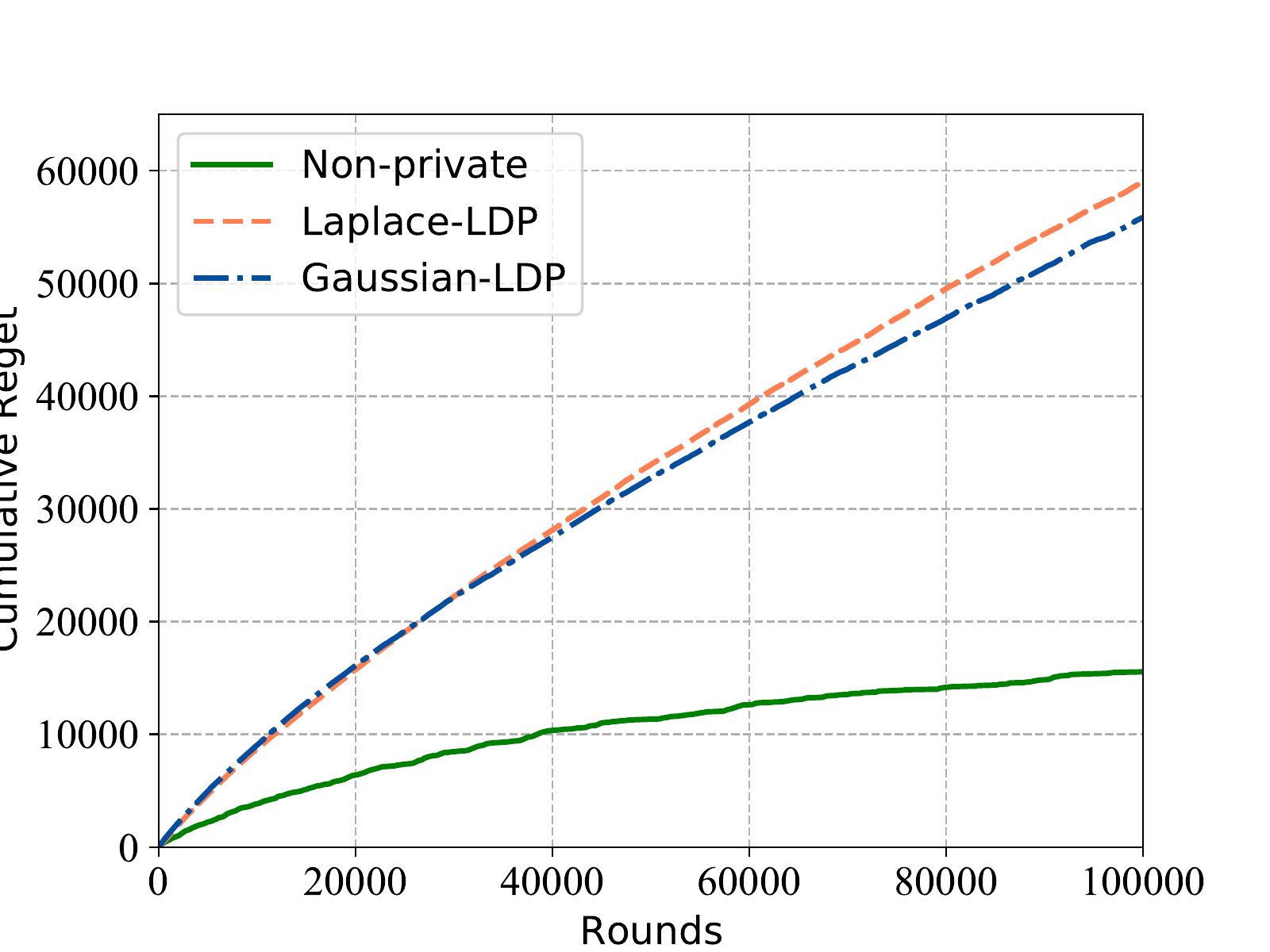}
}
\subfigure[CUCB under LDP, $\epsilon=2$]{ \label{cucb2}
\includegraphics[width=0.31\textwidth]{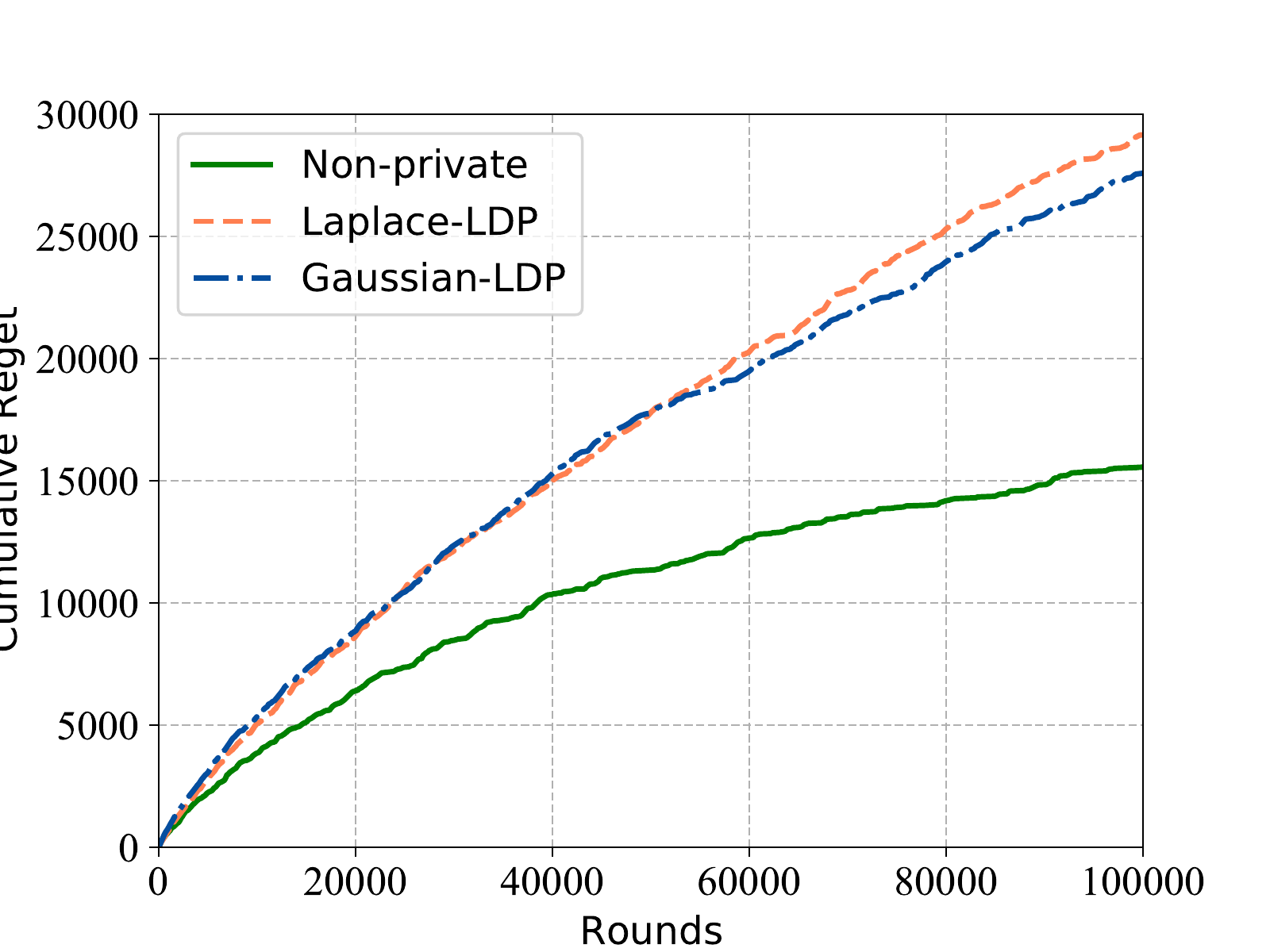}
}
\caption{Empirical results for private and non private algorithms under LDP and DP. }
\end{figure}

\paragraph{Cascading bandits under local differential privacy.} 
The algorithms were analysed empirically under LDP with varying level of $\epsilon$. Figure \ref{ldp1} \ref{ldp2}, \ref{ldp3}, \ref{ldp4} shows the cumulative regret incurred by three algorithms mentioned with $\epsilon = \{0.2, 0.5, 1, 2\}$. Laplace based-UCB achieves the most cumulative regret among all algorithms analyzed. Though non-private UCB incurred the lowest cumulative regret, the private algorithms achieve regrets of order $\mathcal{O}(\log T)$ as well. Figure \ref{ldp5} shows the performance of private algorithms varies according to various number of $\epsilon$ in range of 0 to 2, where $\epsilon$ takes values of every 0.02 interval. When $\epsilon = 0$, the regret is unaffordable for private algorithm. However, as $\epsilon$ increases, the cumulative regret of the algorithms quickly decays and soon achieve comparable cumulative regret with non private algorithm, especially in the case of where Gaussian mechanism is applied. We also test CUCB under $\epsilon = 0.2$ and $\epsilon = 2$ in figure \ref{cucb1}, \ref{cucb2}. It can be observed that, when $\epsilon$ is set to relatively small value, the gap between performance of private and non private algorithm is comparably larger when $\epsilon$ is large. 

\paragraph{Cascading bandits under differential privacy.} 
We provide two sets of empirical results for cascading bandits under DP. Figure \ref{dp1} validates the algorithms under $\epsilon = \{0.2, 0.5, 1, 2\}$. When $\epsilon$ is taken to be a relatively larger value, the performance of private algorithm is closer to the performance of non-private algorithm. Figure \ref{dp2} validates the private algorithms under varying number of arms, for $L = \{8, 12, 16, 20\}$. It can be observed that larger $L$ leads to larger regret.


\section{Conclusion and future work}
\label{section 8}
In this paper, we study the differential privacy and local differential privacy in the cascading bandit setting. Under DP, we give the state-of-the-art $\mathcal{O}(\log^{1+\xi} T)$ regret in the common definition of DP. Under LDP, we utilize the tradeoff between $\epsilon$ and $\delta$, relaxing the $\mathcal{O}(K^2)$ dependence to $\mathcal{O}(K)$. Two respective lower bounds for both situations are provided. We conjecture the algorithm under DP is not good enough to achieve the best regret. For example, during the allocation of privacy budget, all previous work just equally distributes them. However, distributing privacy budget based on arm's difference may improve regret order. We leave this direction for future work.

\bibliographystyle{plainnat} 
\bibliography{ref}

\begin{thebibliography}{27}
\providecommand{\natexlab}[1]{#1}
\providecommand{\url}[1]{\texttt{#1}}
\expandafter\ifx\csname urlstyle\endcsname\relax
  \providecommand{\doi}[1]{doi: #1}\else
  \providecommand{\doi}{doi: \begingroup \urlstyle{rm}\Url}\fi

\bibitem[Abadi et~al.(2016)Abadi, Chu, Goodfellow, McMahan, Mironov, Talwar,
  and Zhang]{abadi2016deep}
Martin Abadi, Andy Chu, Ian Goodfellow, H~Brendan McMahan, Ilya Mironov, Kunal
  Talwar, and Li~Zhang.
\newblock Deep learning with differential privacy.
\newblock In \emph{Proceedings of the 2016 ACM SIGSAC conference on computer
  and communications security}, 2016.

\bibitem[Abbasi-Yadkori et~al.(2011)Abbasi-Yadkori, P{\'a}l, and
  Szepesv{\'a}ri]{abbasi2011improved}
Yasin Abbasi-Yadkori, D{\'a}vid P{\'a}l, and Csaba Szepesv{\'a}ri.
\newblock Improved algorithms for linear stochastic bandits.
\newblock In \emph{Advances in Neural Information Processing Systems}, 2011.

\bibitem[Auer et~al.(2002{\natexlab{a}})Auer, Cesa-Bianchi, and
  Fischer]{auer2002finite}
Peter Auer, Nicolo Cesa-Bianchi, and Paul Fischer.
\newblock Finite-time analysis of the multiarmed bandit problem.
\newblock \emph{Machine learning}, 47\penalty0 (2):\penalty0 235--256,
  2002{\natexlab{a}}.

\bibitem[Auer et~al.(2002{\natexlab{b}})Auer, Cesa-Bianchi, Freund, and
  Schapire]{auer2002nonstochastic}
Peter Auer, Nicolo Cesa-Bianchi, Yoav Freund, and Robert~E Schapire.
\newblock The nonstochastic multiarmed bandit problem.
\newblock \emph{SIAM journal on computing}, 32\penalty0 (1):\penalty0 48--77,
  2002{\natexlab{b}}.

\bibitem[Basu et~al.(2019)Basu, Dimitrakakis, and Tossou]{basu2019differential}
Debabrota Basu, Christos Dimitrakakis, and Aristide Tossou.
\newblock Differential privacy for multi-armed bandits: What is it and what is
  its cost?
\newblock \emph{arXiv preprint arXiv:1905.12298}, 2019.

\bibitem[Besbes et~al.(2014)Besbes, Gur, and Zeevi]{besbes2014stochastic}
Omar Besbes, Yonatan Gur, and Assaf Zeevi.
\newblock Stochastic multi-armed-bandit problem with non-stationary rewards.
\newblock \emph{Advances in neural information processing systems}, 2014.

\bibitem[Bubeck and Cesa{-}Bianchi(2012)]{DBLP:journals/ftml/BubeckC12}
S{\'{e}}bastien Bubeck and Nicol{\`{o}} Cesa{-}Bianchi.
\newblock Regret analysis of stochastic and nonstochastic multi-armed bandit
  problems.
\newblock \emph{Found. Trends Mach. Learn.}, 5\penalty0 (1):\penalty0 1--122,
  2012.

\bibitem[Chan et~al.(2011)Chan, Shi, and Song]{chan2011private}
T-H~Hubert Chan, Elaine Shi, and Dawn Song.
\newblock Private and continual release of statistics.
\newblock \emph{ACM Transactions on Information and System Security},
  14\penalty0 (3):\penalty0 1--24, 2011.

\bibitem[Chen et~al.(2013)Chen, Wang, and Yuan]{chen2013combinatorial}
Wei Chen, Yajun Wang, and Yang Yuan.
\newblock Combinatorial multi-armed bandit: General framework and applications.
\newblock In \emph{International Conference on Machine Learning}, 2013.

\bibitem[Chen et~al.(2020)Chen, Zheng, Zhou, Yang, Chen, and
  Wang]{chen2020locally}
Xiaoyu Chen, Kai Zheng, Zixin Zhou, Yunchang Yang, Wei Chen, and Liwei Wang.
\newblock (locally) differentially private combinatorial semi-bandits.
\newblock In \emph{International Conference on Machine Learning}, 2020.

\bibitem[Cheung et~al.(2019)Cheung, Tan, and Zhong]{cheung2019thompson}
Wang~Chi Cheung, Vincent Tan, and Zixin Zhong.
\newblock A thompson sampling algorithm for cascading bandits.
\newblock In \emph{The 22nd International Conference on Artificial Intelligence
  and Statistics}, 2019.

\bibitem[Dwork et~al.(2006)Dwork, McSherry, Nissim, and
  Smith]{dwork2006calibrating}
Cynthia Dwork, Frank McSherry, Kobbi Nissim, and Adam Smith.
\newblock Calibrating noise to sensitivity in private data analysis.
\newblock In \emph{Theory of cryptography conference}, 2006.

\bibitem[Dwork et~al.(2010)Dwork, Naor, Pitassi, and
  Rothblum]{dwork2010differential}
Cynthia Dwork, Moni Naor, Toniann Pitassi, and Guy~N Rothblum.
\newblock Differential privacy under continual observation.
\newblock In \emph{Proceedings of the forty-second ACM symposium on Theory of
  computing}, 2010.

\bibitem[Dwork et~al.(2014)Dwork, Roth, et~al.]{dwork2014algorithmic}
Cynthia Dwork, Aaron Roth, et~al.
\newblock The algorithmic foundations of differential privacy.
\newblock \emph{Foundations and Trends in Theoretical Computer Science},
  9\penalty0 (3-4):\penalty0 211--407, 2014.

\bibitem[Kairouz et~al.(2015)Kairouz, Oh, and
  Viswanath]{kairouz2015composition}
Peter Kairouz, Sewoong Oh, and Pramod Viswanath.
\newblock The composition theorem for differential privacy.
\newblock In \emph{International conference on machine learning}, 2015.

\bibitem[Kveton et~al.(2015{\natexlab{a}})Kveton, Szepesvari, Wen, and
  Ashkan]{kveton2015cascading}
Branislav Kveton, Csaba Szepesvari, Zheng Wen, and Azin Ashkan.
\newblock Cascading bandits: Learning to rank in the cascade model.
\newblock In \emph{International Conference on Machine Learning},
  2015{\natexlab{a}}.

\bibitem[Kveton et~al.(2015{\natexlab{b}})Kveton, Wen, Ashkan, and
  Szepesvari]{NIPS2015_1f50893f}
Branislav Kveton, Zheng Wen, Azin Ashkan, and Csaba Szepesvari.
\newblock Combinatorial cascading bandits.
\newblock In \emph{Advances in Neural Information Processing Systems},
  2015{\natexlab{b}}.

\bibitem[Li et~al.(2016)Li, Wang, Zhang, and Chen]{li2016contextual}
Shuai Li, Baoxiang Wang, Shengyu Zhang, and Wei Chen.
\newblock Contextual combinatorial cascading bandits.
\newblock In \emph{International conference on machine learning}, 2016.

\bibitem[Mishra and Thakurta(2015)]{mishra2015nearly}
Nikita Mishra and Abhradeep Thakurta.
\newblock (nearly) optimal differentially private stochastic multi-arm bandits.
\newblock In \emph{Proceedings of the Thirty-First Conference on Uncertainty in
  Artificial Intelligence}, 2015.

\bibitem[Ren et~al.(2020)Ren, Zhou, Liu, and Shroff]{ren2020multi}
Wenbo Ren, Xingyu Zhou, Jia Liu, and Ness~B Shroff.
\newblock Multi-armed bandits with local differential privacy.
\newblock \emph{arXiv preprint arXiv:2007.03121}, 2020.

\bibitem[Rigollet and H{\"u}tter(2015)]{rigollet2015high}
Phillippe Rigollet and Jan-Christian H{\"u}tter.
\newblock High dimensional statistics.
\newblock \emph{Lecture notes for course 18S997}, 813:\penalty0 814, 2015.

\bibitem[Shariff and Sheffet(2018)]{NEURIPS2018_a1d7311f}
Roshan Shariff and Or~Sheffet.
\newblock Differentially private contextual linear bandits.
\newblock In \emph{Advances in Neural Information Processing Systems}, 2018.

\bibitem[Tossou and Dimitrakakis(2016)]{tossou2016algorithms}
Aristide Tossou and Christos Dimitrakakis.
\newblock Algorithms for differentially private multi-armed bandits.
\newblock In \emph{Proceedings of the AAAI Conference on Artificial
  Intelligence}, 2016.

\bibitem[Tossou and Dimitrakakis(2017)]{tossou2017achieving}
Aristide Tossou and Christos Dimitrakakis.
\newblock Achieving privacy in the adversarial multi-armed bandit.
\newblock In \emph{Proceedings of the AAAI Conference on Artificial
  Intelligence}, 2017.

\bibitem[Wang et~al.(2021)Wang, Zhao, Li, and Shao]{wang2021conservative}
Kun Wang, Canzhe Zhao, Shuai Li, and Shuo Shao.
\newblock Conservative contextual combinatorial cascading bandit.
\newblock \emph{arXiv preprint arXiv:2104.08615}, 2021.

\bibitem[Zhao et~al.(2019)Zhao, Wang, Bai, Lam, Xu, Shi, Ren, Yang, Liu, and
  Yu]{zhao2019reviewing}
Jun Zhao, Teng Wang, Tao Bai, Kwok-Yan Lam, Zhiying Xu, Shuyu Shi, Xuebin Ren,
  Xinyu Yang, Yang Liu, and Han Yu.
\newblock Reviewing and improving the gaussian mechanism for differential
  privacy.
\newblock \emph{arXiv preprint arXiv:1911.12060}, 2019.

\bibitem[Zong et~al.(2016)Zong, Ni, Sung, Ke, Wen, and
  Kveton]{DBLP:conf/uai/ZongNSKWK16}
Shi Zong, Hao Ni, Kenny Sung, Nan~Rosemary Ke, Zheng Wen, and Branislav Kveton.
\newblock Cascading bandits for large-scale recommendation problems.
\newblock In \emph{Proceedings of the Thirty-Second Conference on Uncertainty
  in Artificial Intelligence}, 2016.

\end{thebibliography}

\newpage
\appendix

\section{Technical Lemmas}

\begin{lemma}[\citet{kveton2015cascading}]
\label{lem_4}
For any item $e$ and optimal item $e^{\ast}$, let:
$G_{e,e^{\ast},t}=\{\exists 1\leq k \leq K \quad s.t.\quad a_k^t=e, \pi_t(k)=e^{\ast},w_t(a_1^t)=...=w_t(a_{k-1}^t)=0\}$, then there exists a permutation $\pi_t$ of optimal items $\{1, ..., K\}$, which is a deterministic function of $\mathcal{H}_t$, such that $U_t(a^t_k) \geq U_t(\pi_t(k))$ for all $k$ and the following holds:
\begin{align*}
    \mathbb{E}_t[R(A_t,w_t)] &\geq \alpha \sum_{e=K+1}^L \sum_{e^{\ast}=1}^K \Delta_{e,e^{\ast}} \mathbb{E}_t[\mathbbm{1}\{G_{e,e^{\ast},t}\}]\,,\\
    \mathbb{E}_t[R(A_t,w_t)] &\leq  \sum_{e=K+1}^L \sum_{e^{\ast}=1}^K \Delta_{e,e^{\ast}} \mathbb{E}_t[\mathbbm{1}\{G_{e,e^{\ast},t}\}]\,,
\end{align*}
where $\alpha=(1-\bar{w}(1))^{K-1}$ and $\bar{w}(1)$ is the attraction probability of the most attractive item. 

\end{lemma}

\begin{lemma}[\cite{chan2011private}]
\label{llem6}
Let $X_1,X_2,...X_n$ be i.i.d. random variables following the Lap(b) distribution and $Y=X_1+X_2+...+X_n$. For $v \geq b\sqrt{n}$ and $0<\lambda < \frac{2\sqrt{2}v^2}{b}$, we have $$P\{ Y > \lambda \} \leq \exp(-\frac{\lambda^2}{8v^2})\,.$$
\end{lemma}

\begin{lemma}[\citet{rigollet2015high}]
\label{lem4}
If $X_1,...X_n$ are i.i.d $\mathcal{N}(u,\sigma^2)$. $\bar{X}=\frac{1}{n} \sum_{i=1}^n X_i \sim \mathcal{N}(u,\sigma^2/n)$. The following inequality holds:
$$P(|\bar{X}-u|>t) \leq 2\exp(-\frac{nt^2}{2\sigma^2})\,.$$
\end{lemma}

\section{CUCB under LDP}

	\begin{algorithm}[H]
	\label{algorithm 4}
		\caption{CUCB under LDP (Gaussian Mechanism).}
		\KwIn{$\epsilon$: the differential privacy parameter, $m$: the number of arm, $K$: the maximum arm each round recommends. }
		Initialization:$\sigma=\frac{1}{\epsilon}\sqrt{2K \ln \frac{1.25}{\delta}},\gamma=2t^{-2}$\\
		Observe $u_0$.\\
		$\forall i \in [m]:T_i \leftarrow 1$\\
		$\forall i \in [m]:\hat{u}(i)\leftarrow u_0(i)$\\
		\For{$t =m+1,m+2,...,n$}{
		    $\forall i \in [m]:\bar{u}_i=\hat{u}_i+\sqrt{\frac{3 \log t}{2T_i}}+\frac{2}{\epsilon}\sqrt{\frac{2K \log \frac{1.25}{\delta}\log t}{T_i}}$\\
		    $S_t=\text{Oracle}(\bar{u}_1,\bar{u}_2,...,\bar{u}_K)$.\\
		    Play S and observe $u_t$.\\
		    \For{$e=1,...,K$}{
		        $T_{t}(e)=T_{t-1}(e)+1.$\\
		        $\hat{u}_{T_t(e)}(e)=\frac{T_{t-1}(e)\hat{u}_{T_{t-1}(e)}(e)+u_{t}(e)+\mathcal{N}(0,\sigma^2) }{T_t(e)}$\\
		    }
		}
	\end{algorithm}

		    

	
		    


\section{Proof of Theorem 2}
\begin{proof}
According to Lemma \ref{lem_4}, the instant regret at time $t$ satisfies following inequality:
$$ R_t \leq \sum_{e=K+1}^L \sum_{e^{\ast}=1}^K\Delta_{e,e^{\ast}}\mathbb{E}_t[\mathbbm{1}\{G_{e,e^{\ast},t}\}]\,.$$

Let $\tilde{w}$ denote the empirical mean not adding Laplace noise,  $\Lambda_{1,t}=\{\forall e\in E , \text{ s.t. } |\bar{w}(e)-\tilde{w}(e)| \leq c_{t,T_{t-1}(e)} \}$,
$\Lambda_{2,t}=\{\forall e\in E , \text{ s.t. } |\frac{Noise_{t,T_{t-1}(e)}}{T_{t-1}(e)}| \leq v_{t,T_{t-1}(e)}\}$, and $T_{t-1}(e)$ denotes the number of arm $e$ pulled until time $t-1$,

$$\mathbb{E}[\sum_{t=1}^T \mathbbm{1}\{ \bar{\Lambda}_{1,t} \}]\leq \sum_{e \in E}\sum_{t=1}^T \sum_{s=1}^t P(|\bar{w}(e)-\tilde{w}(e)| \geq c_{t,s}) \leq 2 \sum_{e \in E}\sum_{t=1}^T \sum_{s=1}^t e^{-3 \log t} \leq 2 \sum_{e \in E} \sum_{t=1}^T t^{-2}\leq \frac{\pi^2}{3}L\,.$$

Based on Lemma \ref{lem1}, let $\gamma=t^{-3}$, then $v_{t,T_{t-1}(e)}=\frac{3c_1L\log^{1.5}T_{t-1}(e)\log t}{\epsilon T_{t-1}(e)}$,
$$\mathbb{E}[\sum_{t=1}^T \mathbbm{1}\{ \bar{\Lambda}_{2,t} \}]\leq \sum_{e \in E}\sum_{t=1}^T \sum_{s=1}^t P(\left|\frac{Noise_{t,s}}{s}\right| \geq v_{t,s}) \leq  \sum_{e \in E}\sum_{t=1}^T \sum_{s=1}^t t^{-3} \leq  \sum_{e \in E} \sum_{t=1}^T t^{-2} < \frac{\pi^2}{3}L\,.$$
$$R(T)\leq \mathbb{E}[\sum_{t=1}^T \mathbbm{1}\{\Lambda_{1,t},\Lambda_{2,t}\}R_t]+\mathbb{E}[\sum_{t=1}^T\mathbbm{1}\{\bar{\Lambda}_{1,t}\}R_t]+\mathbb{E}[\sum_{t=1}^T\mathbbm{1}\{\bar{\Lambda}_{2,t}\}R_t]\,.$$

Then since $R_t \leq 1$, the last two terms can be bounded by $\frac{2\pi^2}{3}L$, and for all time $t$ with probability of $1-t^{-3}$,
$$\left|\frac{Noise_{t,T_{t-1}(e)}}{T_{t-1}(e)}\right| \leq \frac{3c_1L\log^{1.5}T_{t-1}(e)\log t}{\epsilon T_{t-1}(e)}=v_{t,T_{t-1}(e)}\,.$$

Therefore, the cumulative regret until time $T$ can be bounded as following:
$$R(T)\leq  \sum_{e=K+1}^L \mathbb{E}[\sum_{e^{\ast}=1}^K \sum_{t=1}^T \Delta_{e,e^{\ast}} \mathbbm{1}\{\Lambda_{1,t}, \Lambda_{2,t}, G_{e,e^{\ast},t}\}]+\frac{2 \pi^2L}{3}\,.$$

If at time t, the algorithm selects a sup-optimal arm, then 
$$\bar{w}(e)+2 c_{t-1,T_{t-1}(e)} + 2v_{t-1,T_{t-1}(e)} \geq U_t(e) \geq U_t(e^{\ast}) \geq \bar{w}(e^{\ast})\,.$$
$$\Delta_{e,e^{\ast}} \leq 2c_{t-1,T_{t-1}(e)}+2v_{t-1,T_{t-1}(e)}\,.$$
We consider its complementary set, i.e. $\mathcal{E}_{t}=\{\Delta_{e,e^{\ast}} > 2c_{t-1,T_{t-1}(e)}+2v_{t-1,T_{t-1}(e)}\}$. It indicates the event the algorithm always chooses the optimal arm. It is enough to bound the following two inequalities:
$$\lambda_0\Delta> 2c_{t,T_{t-1}(e)}\,,$$
$$(1-\lambda_0) \Delta > 2v_{t-1,T_{t-1}(e)}\,,$$
For the first inequality, we get:
$$T_{t-1}(e) \geq \frac{6}{\lambda_0^2\Delta_{e,e^{\ast}}^2}\log t\,.$$

For the second inequality, we get:
$$T_{t-1}(e) \geq B \log^{1.5} T_{t-1}(e)\,,$$
where $B=\frac{6c_1L\log t}{\epsilon (1-\lambda)\Delta}$.\\
If we let $T_{t-1}(e)=B^{1+\xi}$, we get
$$B^{\xi} \geq (1+\xi)^{1.5} \log^{1.5} B\,.$$
When $t$ is sufficiently large, it must exist a constant $c_2$, when $B>c_2$, the above inequality holds. Thus,
$$T_{t-1}(e) \geq \max \{\frac{6}{\lambda_0^2\Delta_{e,e^{\ast}}^2}\log t,B^{1+\xi}\}\,.$$
It means that, when $T_{t-1}(e) \geq \max \{\frac{6}{\lambda_0^2\Delta_{e,e^{\ast}}^2}\log t,B^{1+\xi}\}$, the algorithm must choose the optimal arm at this time. What's more, we consider $t$ as the dominant term, and return back to previous situation,
$$\Delta_{e,e^{\ast}} \leq 2c_{t-1,T_{t-1}(e)}+2v_{t-1,T_{t-1}(e)} \Rightarrow T_{t-1}(e) \leq B^{1+\xi}=\tau_{e,e^{\ast}}\,.$$

Therefore,
\begin{equation}
\label{equa1}
\begin{aligned}
    \sum_{e^{\ast}=1}^K \sum_{t=1}^T \Delta_{e,e^{\ast}} \mathbbm{1}\{\Lambda_{1,t}, \Lambda_{2,t}, G_{e,e^{\ast},t}\} \leq \sum_{e^{\ast}=1}^K \Delta_{e,e^{\ast}} \sum_{t=1}^T \mathbbm{1}\{T_{t-1}(e) \leq \tau_{e,e^{\ast}},G_{e,e^{\ast},t}\}\,.
\end{aligned}
\end{equation}

Let 
\[
M_{e,e^\ast}=\sum_{t=1}^T \mathbbm{1}\{ T_{t-1}(e) \leq \tau_{e,e^\ast},G_{e,e^\ast,t}\}.
\]

Now note that (i) the counter ${T}_{t - 1}(e)$ of item $e$ increases by one when the event $G_{e, e^\ast, t}$ happens for any optimal item $e^\ast$, (ii) the event $G_{e, e^\ast, t}$ happens for at most one optimal $e^\ast$ at any time $t$; and (iii) $\tau_{e, 1} \leq ... \leq \tau_{e, K}$. Based on these facts, it follows that ${M}_{e, e^\ast} \leq \tau_{e, e^\ast}$, and moreover $\sum_{e^\ast = 1}^K {M}_{e, e^\ast} \leq \tau_{e, K}$. Therefore, the right-hand side of (\ref{equa1}) can be bounded by:
$$\max\left\{\sum_{e^{\ast}=1}^K \Delta_{e,e^{\ast}} m_{e,e^{\ast}}:0\leq m_{e,e^{\ast}} \leq \tau_{e,e^{\ast}}, \sum_{e^{\ast}=1}^K m_{e,e^{\ast}} \leq \tau_{e,K}\right\}\,.$$
Since the gaps are decreasing, $\Delta_{e, 1} \geq \ldots \geq \Delta_{e, K}$, the solution to the above problem is $m_{e, 1}^\ast = \tau_{e, 1}$, $m_{e, 2}^\ast = \tau_{e, 2} - \tau_{e, 1}$, $\dots$, $m_{e, K}^\ast = \tau_{e, K} - \tau_{e, K - 1}$. Therefore, the above can be bounded by:
$$\left[\Delta_{e,1}\frac{1}{\Delta_{e,1}^{1+\xi}}+\sum_{e^{\ast}=2}^K \Delta_{e,e^{\ast}}(\frac{1}{\Delta_{e,e^{\ast}}^{1+\xi}}-\frac{1}{\Delta_{e,e^{\ast}-1}^{1+\xi}})\right](\frac{\sqrt{8}c_1L}{\epsilon(1-\lambda_0)} \ln 4T^2)^{1+\xi}\,.$$

Therefore, let $\lambda_0=1/2$, we get:
$$\sum_{e^{\ast}=1}^K \sum_{t=1}^T \Delta_{e,e^{\ast}} \mathbbm{1}\{\Lambda_{1,t}, \Lambda_{2,t}, G_{e,e^{\ast},t}\} \leq \frac{192c_1^{1+\xi}L^{1+\xi}}{\Delta_{e,K}^{\xi} \epsilon^{1+\xi}} \log^{1+\xi} T\,.$$
\[
    R(T)\leq \sum_{e=K+1}^L \frac{192c_1^{1+\xi} L^{1+\xi}}{\Delta_{e,K}\epsilon^{1+\xi}}\log^{1+\xi} T +\frac{2\pi^2}{3}L+c_2\,. \qedhere
\]
\end{proof}


\section{Proof of Theorem 4}
\begin{proof}

Let $\tilde{w}$ denote the empirical mean not adding differential private noise, $\Lambda_{1,t}=\{\forall e\in E \text{ s.t. } |\bar{w}(e)-\tilde{w}(e)| \leq c_{t,T_{t-1}(e)} \}$,
$\Lambda_{2,t}=\{\forall e\in E \text{ s.t. } |\frac{\sum_{t=1}^{T_{t-1}(e)} Lap}{T_{t-1}(e)}| \leq \frac{K}{\epsilon}\sqrt{\frac{24\log t}{T_{t-1}(e)}}\}$, the cumulative regret until time $t$ is:
$$R(T)\leq \mathbb{E}[\sum_{t=1}^T \mathbbm{1}\{\Lambda_{1,t},\Lambda_{2,t}\}R_t]+\mathbb{E}[\sum_{t=1}^T\mathbbm{1}\{\bar{\Lambda}_{1,t}\}R_t]+\mathbb{E}[\sum_{t=1}^T\mathbbm{1}\{\bar{\Lambda}_{2,t}\}R_t]\,.$$
$$\mathbb{E}[\sum_{t=1}^T \mathbbm{1}\{ \bar{\Lambda}_{1,t} \}]\leq \sum_{e \in E}\sum_{t=1}^T \sum_{s=1}^t P(|\bar{w}(e)-\tilde{w}(e)| \geq c_{t,s}) \leq 2 \sum_{e \in E}\sum_{t=1}^T \sum_{s=1}^t e^{-3 \log t} \leq 2 \sum_{e \in E} \sum_{t=1}^T t^{-2}\leq \frac{\pi^2}{3}L\,.$$

According to Lemma \ref{llem6}, let $\lambda=\frac{K}{\epsilon}\sqrt{24s \log t}$,$b=\frac{K}{\epsilon}$,$v=\frac{K\sqrt{s}}{\epsilon}$,$\lambda< \frac{2\sqrt{2}Ks}{\epsilon}$, when $s \geq 3\log t$ (this part can be bounded by $3\log t$),
$$\mathbb{E}[\sum_{t=1}^T \mathbbm{1}\{ \bar{\Lambda}_{2,t} \}]\leq \sum_{e \in E}\sum_{t=1}^T \sum_{s=1}^t 2 P(\left|\frac{\sum_{t=1}^s Lap}{s}\right| \geq \frac{K}{\epsilon}\sqrt{\frac{24 \log t}{s}}) \leq 2 \sum_{e \in E}\sum_{t=1}^T \sum_{s=1}^t t^{-3} \leq 2 \sum_{e \in E} \sum_{t=1}^T t^{-2}\leq \frac{\pi^2}{3}L\,,$$
Then based on Lemma \ref{lem_4}, we get:
$$\mathbb{E}[\sum_{t=1}^T \mathbbm{1}\{\Lambda_{1,t},\Lambda_{2,t}\}R_t]\leq \sum_{e=K+1}^L \Delta_{e,e^{\ast}} \mathbb{E}[\sum_{e^{\ast}=1}^{K} \sum_{t=1}^T \mathbbm{1}\{\Lambda_{1,t},\Lambda_{2,t},G_{e,e^{\ast},t}\}]\,.$$

Now we bound the above inequality for any sub-optimal arms, let $h_{t-1,s}= \frac{K}{\epsilon}\sqrt{\frac{24 \log t}{s}}$, if at time $t$, the algorithm selects a sub-optimal arm e, then:
$$\bar{w}(e)+2 c_{t-1,T_{t-1}(e)} + 2h_{t-1,T_{t-1}(e)} \geq U_t(e) \geq U_t(e^{\ast}) \geq \bar{w}(e^{\ast})\,,$$
which implies:
$$\Delta_{e,e^{\ast}} \leq 2c_{t-1,T_{t-1}(e)}+2h_{t-1,T_{t-1}(e)}\,.$$

Then for any sub-optimal arm $e$, we have:
$$T_{t-1}(e) \leq \frac{4 \log t(\sqrt{1.5}+\frac{K}{\epsilon}\sqrt{24})^2}{\Delta_{e,e^{\ast}}^2}=\tau_{e,e^{\ast}}\,.$$
Therefore,
\begin{equation}
\label{equa2}
\begin{aligned}
    \sum_{e^{\ast}=1}^K \sum_{t=1}^T \Delta_{e,e^{\ast}} \mathbbm{1}\{\Lambda_{1,t}, \Lambda_{2,t}, G_{e,e^{\ast},t}\} \leq \sum_{e^{\ast}=1}^K \Delta_{e,e^{\ast}} \sum_{t=1}^T \mathbbm{1}\{T_{t-1}(e) \leq \tau_{e,e^{\ast}},G_{e,e^{\ast},t}\}\,.
\end{aligned}
\end{equation}

Let 
\[
M_{e,e^\ast}=\sum_{t=1}^T \mathbbm{1}\{ T_{t-1}(e) \leq \tau_{e,e^\ast},G_{e,e^\ast,t}\}
\]

Now note that (i) the counter ${T}_{t - 1}(e)$ of item $e$ increases by one when the event $G_{e, e^\ast, t}$ happens for any optimal item $e^\ast$, (ii) the event $G_{e, e^\ast, t}$ happens for at most one optimal $e^\ast$ at any time $t$; and (iii) $\tau_{e, 1} \leq ... \leq \tau_{e, K}$. Based on these facts, it follows that ${M}_{e, e^\ast} \leq \tau_{e, e^\ast}$, and moreover $\sum_{e^\ast = 1}^K {M}_{e, e^\ast} \leq \tau_{e, K}$. Therefore, the right-hand side of (\ref{equa2}) can be bounded by:
$$\max\left\{\sum_{e^{\ast}=1}^K \Delta_{e,e^{\ast}} m_{e,e^{\ast}}:0\leq m_{e,e^{\ast}} \leq \tau_{e,e^{\ast}}, \sum_{e^{\ast}=1}^K m_{e,e^{\ast}} \leq \tau_{e,K}\right\}\,.$$
Since the gaps are decreasing, $\Delta_{e, 1} \geq \ldots \geq \Delta_{e, K}$, the solution to the above problem is $m_{e, 1}^\ast = \tau_{e, 1}$, $m_{e, 2}^\ast = \tau_{e, 2} - \tau_{e, 1}$, $\dots$, $m_{e, K}^\ast = \tau_{e, K} - \tau_{e, K - 1}$. Therefore, the above can be bounded by:
\begin{align*}
  \left[\Delta_{e, 1} \frac{1}{\Delta_{e, 1}^2} + \sum_{e^\ast = 2}^K \Delta_{e, e^\ast}
  \left(\frac{1}{\Delta_{e, e^\ast}^2} - \frac{1}{\Delta_{e, e^\ast - 1}^2}\right)\right]4(\sqrt{1.5}+K/\epsilon\sqrt{24})^2  \log T\,.
\end{align*}

The above term is bounded by
$$\frac{8(\sqrt{1.5}+\frac{K}{\epsilon}\sqrt{24})^2}{\Delta_{e,K} }\log T\,.$$
Therefore,
\[R(T)\leq \sum_{e=K+1}^L \frac{8(\sqrt{1.5}+\frac{K}{\epsilon}\sqrt{24})^2}{\Delta_{e,K} }\log T+\frac{2\pi^2}{3}L =\mathcal{O}\left(\sum_{e=K+1}^{L} \frac{K^2}{\epsilon^2\Delta_{e,K}} \log T\right)\,. \qedhere
\]

\end{proof}

\section{Proof of Theorem 6}

\begin{proof}

Following the same procedure as Appendix D, let $\tilde{w}$ denote the empirical mean not adding differential private noise, $\Lambda_{1,t}=\{\forall e\in E \text{ s.t. } |\bar{w}(e)-\tilde{w}(e)| \leq c_{t,T_{t-1}(e)} \}$,
$\Lambda_{2,t}=\{\forall e\in E \text{ s.t. } | \frac{\sum_{t=1}^{T_{t-1}(e)} N_t}{T_{t-1}(e)}| \leq \frac{2}{\epsilon}\sqrt{\frac{K \log \frac{1.25}{\delta}\log \frac{2}{\gamma}}{T_{t-1}(e)}}\}$. 

$$R(T)\leq \mathbb{E}[\sum_{t=1}^T \mathbbm{1}\{\Lambda_{1,t},\Lambda_{2,t}\}R_t]+\mathbb{E}[\sum_{t=1}^T \mathbbm{1}\{\bar{\Lambda}_{1,t}\}R_t]+\mathbb{E}[\sum_{t=1}^T \mathbbm{1}\{\bar{\Lambda}_{2,t}\}R_t]\,.$$

Based on Lemma \ref{lem4}, we get with probability at least $1-\gamma$,
$u \in [\bar{X}-\frac{2}{\epsilon}\sqrt{\frac{K \log \frac{1.25}{\delta}\log \frac{2}{\gamma}}{T_{t-1}(e)}},\bar{X}+\frac{2}{\epsilon}\sqrt{\frac{K \log \frac{1.25}{\delta}\log \frac{2}{\gamma}}{T_{t-1}(e)}}]$. Let error probability for Gaussian distribution $\gamma=t^{-3}$, then the latter two terms can be bounded by $\frac{2\pi^2}{3}L$. As for the first term,

$$\mathbb{E}[\sum_{t=1}^T \mathbbm{1}\{\Lambda_{1,t},\Lambda_{2,t}\}R_t]\leq \sum_{e=K+1}^L \Delta_{e,e^{\ast}} \mathbb{E}[\sum_{e^{\ast}=1}^{K} \sum_{t=1}^T \mathbbm{1}\{\Lambda_{1,t},\Lambda_{2,t},G_{e,e^{\ast},t}\}]\,.$$

Let $h_{t-1,s}=\frac{2}{\epsilon}\sqrt{\frac{K \log \frac{1.25}{\delta}\log 2t^3}{s}}$, if the algorithm chooses a sub-optimal arm at round $t$,

$$\bar{w}(e)+2 c_{t-1,T_{t-1}(e)} + 2h_{t-1,T_{t-1}(e)} \geq U_t(e) \geq U_t(e^{\ast}) \geq \bar{w}(e^{\ast})\,,$$
$$2c_{t-1,T_{t-1}(e)}+2h_{t-1,T_{t-1}(e)} \geq \Delta_{e,e^{\ast}}\,.$$
$$2\sqrt{\frac{1.5 \log t}{T_{t-1}(e)}}+\frac{4}{\epsilon}\sqrt{\frac{K \log \frac{1.25}{\delta}\log 2t^3}{T_{t-1}(e)}} \geq \Delta_{e,e^{\ast}}\,.$$
Then we get,
$$T_{t-1}(e) \leq \frac{(2\sqrt{1.5}+\frac{8}{\epsilon}\sqrt{K\log \frac{1.25}{\delta}})^2\log t}{\Delta_{e,e^{\ast}}^2}\,.$$
Follow the same procedure as Appendix D, finally we get:
\[
R(T) \leq \sum_{e=K+1}^{L}\frac{2 (2\sqrt{1.5}+\frac{8}{\epsilon}\sqrt{K\log \frac{1.25}{\delta}})^2}{\Delta_{e,K}}\log T+\frac{2\pi^2}{3}L=\mathcal{O}\left(\sum_{e=K+1}^{L}\frac{K\log 1/\delta}{\epsilon^2\Delta_{e,K}}\log T \right)\,.     \qedhere  
\]
\end{proof}

\section{Proof of Theorem 9}

\begin{proof}

Before the proof, we first define some notations based on \cite{chen2013combinatorial}: let $T_{i,t}$ be the number of arm $i$ pulled at the end of time $t$. The oracle in the algorithm is a $(\alpha,\beta)$-approximation oracle, which means  $P[r_u(S) \geq \alpha \text{opt}_u] \geq \beta$. A super arm $S$ is bad if $r_u(S)<\alpha \text{opt}_u$. And let $S_B=\{S|r_u(S)< \alpha \text{opt}_u\}$. Under this circumstance, 
$$
\Delta_{min}^i=\alpha \text{opt}_u-\max\{r_u(S)|S \in S_B, i \in S\},$$
$$\Delta_{max}^i=\alpha \text{opt}_u-\min\{r_u(S)|S \in S_B, i \in S\}.\\
$$
Following above, the cumulative regret has the following form:
\[
R(T)=T\alpha \beta \text{opt}_u-\mathbb{E}[\sum_{i=1}^T r_u(S_t)]
\]

Round $t$ is bad if the oracle selects a bad super arm $S_t \in S_B$.  Let $F_t$ be the event that the oracle fails to produce an $\alpha$-approximate answer. We have $P[F_t]\leq 1-\beta$. Moreover, we maintain counter $N_{i}$ for each arm $i$ after $m$ initialization rounds. If at round $t\geq m$, the oracle selects a bad super arm, let $i=\arg \min_{j \in S_t} N_{j,t-1}$, then $N_{i,t}=N_{i,t-1}+1$. By definition, $N_{i,t} \leq T_{i,t}$. Note that in every bad round, exactly one counter in $\{N_i\}_{i=1}^m$ is incremented, so the total number of bad rounds in the first $T$ rounds is less than or equal $\sum_i N_{i,T}$.

Let $l_t=\frac{128K\log \frac{1.25}{\delta} \log t }{\min\{\epsilon^2,2\}(f^{-1}(\Delta_{min}))^2}$. Consider a bad round $t$, i.e. $S_t \in S_B$ is selected at time $t$, then we have:

\begin{equation}
    \begin{aligned}
    &\quad\sum_{i=1}^m N_{i,T} -m(l_T+1)\\
    &=\sum_{t=m+1}^T \mathbbm{1}\{S_t \in S_B\}-ml_T\\
    &\leq \sum_{t=m+1}^T \sum_{i\in[m]} \mathbbm{1}\{S_t \in S_B,N_{i,t} >N_{i,t-1}, N_{i,t-1}>l_T\}\\
    &\leq \sum_{t=m+1}^T \sum_{i\in[m]} \mathbbm{1}\{S_t \in S_B,N_{i,t} >N_{i,t-1}, N_{i,t-1}>l_t\}\\
    &=\sum_{t=m+1}^T \mathbbm{1}\{S_t \in S_B, \forall i \in S_t, N_{i,t-1}> l_t \}\\
    &\leq\sum_{t=m+1}^T(\mathbbm{1}\{F_t\}+\mathbbm{1}\{\lnot F_t,S_t \in S_B, \forall i \in S_t, N_{i,t-1}>l_t\})\\
        &\leq\sum_{t=m+1}^T(\mathbbm{1}\{F_t\}+\mathbbm{1}\{\lnot F_t,S_t \in S_B, \forall i \in S_t, T_{i,t-1}>l_t\})\,.\\
    \end{aligned}
\end{equation}
Let $\tilde{u}$ denote the empirical mean not adding differential private noise, and $G_t=\{\forall i \in [m], \left|\frac{\sum_t N_t}{T_i}\right| \leq \frac{2}{\epsilon}\sqrt{\frac{2K \log \frac{1.25}{\delta}\log t}{T_i}}\}$,
\begin{equation}
    \begin{aligned}
    &P[|\tilde{u}_i-u_i|\geq \sqrt{3\log t/(2T_{i,t-1})}]\\
    &\leq \sum_{s=1}^{t-1} Pr[\{|\hat{u}-u_i|\geqq \sqrt{3\log t/(2s),T_{i,t-1}=s}\}]\\
    &\leq \sum_{s=1}^{t-1} Pr[|\hat{u}_{i,s}-u_i|\geq \sqrt{3 \log t/(2s)}]\\
    &\leq 2te^{-3\log t}=2t^{-2}\,.\\
    \end{aligned}
\end{equation}

Let $\gamma=2t^{-2}$, then with probability of $1-2t^{-2}$,
$$ \left|\frac{\sum_t N_t}{T_i}\right| \leq \frac{2}{\epsilon}\sqrt{\frac{2K \log \frac{1.25}{\delta}\log t}{T_i}}\,.$$

Let $\Lambda_{i,t}=2\sqrt{\frac{8K\log 1.25/\delta \log t}{\min\{\epsilon^2,2\} T_i}}$, then $\Lambda_{i,t}$ is the max confidence interval each round because:

$$\Lambda_{i,t} \geq \frac{2}{\epsilon}\sqrt{\frac{2K \log \frac{1.25}{\delta}\log t}{T_i}}+\sqrt{\frac{3 \log t}{2 T_i}}\,.$$

Let $E_t=\{\forall i \in [m],|\hat{u}_{i,T_{i,t-1}}-u_i|\leq \Lambda_{i,t}\}$, $\Lambda_t=\max\{\Lambda_{i,t}| i \in S_t\}$, $\Lambda=2\sqrt{\frac{8K\log 1.25/\delta \log t}{\min\{\epsilon^2,2\}l_t}}$ which is not a random variable, we have:
$$P[\lnot E_t]\leq 4mt^{-2}\,,$$

$$E_t \Rightarrow \forall i \in S_t, |\bar{u}_{i,t}-u_i| <2 \Lambda_t\,,$$

$$\{S_t \in S_B, \forall i \in S_t, T_{i,t-1}>l_t\} \Rightarrow \Lambda > \Lambda_t\,,$$

$$E_t \Rightarrow \bar{u}_t \geq u\,.$$

If $\{E_t,\lnot F_t, S_t \in S_B, \forall i \in S_t, T_{i,t-1}>l_t\}$ holds at time $t$, we have the following important derivation:
\[
    r_u(S_t)+f(2\Lambda)>r_u(S_t)+f(2\Lambda_t) \geq r_{\bar{u}_t}(S_t)\geq \alpha \text{opt}_{\bar{u}_t} \geq \alpha r_{\bar{u}_t}(S_t^{\ast}) \geq r_{u}(S_u^{\ast})=\alpha \text{opt}_u \,.
\]
So we have 
$$r_u(S_t)+f(2\Lambda)>\alpha \text{opt}_u\,.$$

Since $l_t=\frac{128K\log \frac{1.25}{\delta} \log t }{\min\{\epsilon^2,2\}(f^{-1}(\Delta_{min}))^2}$, we have $f(2\Lambda)=\Delta_{min}$. This contradicts the definition of $\Delta_{min}$ and the fact $S_t \in S_B$. Thus we have:

\begin{equation}
    \begin{aligned}
    P[\{E_t,\lnot F_t,S_t \in S_B, \forall i \in S_t, T_{i,t-1}>l_t\}]=0 \Rightarrow\\
    P[\{\lnot F_t,S_t \in S_B, \forall i \in S_t, T_{i,t-1}>l_t\}]\leq P[\lnot E_t]\leq 4mt^{-2}\,.\\
    \end{aligned}
\end{equation}
\begin{equation}
    \begin{aligned}
    &\mathbb{E}[\sum_{i=1}^m N_{i,T}] \leq m(l_T+1)+(1-\beta)(T-m)+\sum_{t=1}^T \frac{4m}{t^2}\\
    &\leq \frac{128mK\log \frac{1.25}{\delta} \log T }{\min\{\epsilon^2,2\}(f^{-1}(\Delta_{min}))^2}+(\frac{2\pi^2}{3}+1)m+(1-\beta)(T-m)\,.
    \end{aligned}
\end{equation}
\begin{align*}
    R(T)
    &\leq T\alpha \beta \text{opt}_u-(T\alpha \text{opt}_u-\mathbb{E}\left[\sum_{i=1}^m N_{i,T}\right]\Delta_{max})\\
    &\leq T\alpha(\beta-1)\text{opt}_u+\mathbb{E}\left[\sum_{i=1}^m N_{i,T}\right]\Delta_{max}\\
    &\leq \left[\frac{128K\log \frac{1.25}{\delta} \log T}{\min\{\epsilon^2,2\}(f^{-1}(\Delta_{min}))^2}+\frac{2\pi^2}{3}+1\right]m \Delta_{max}\\
    &=\mathcal{O}\left(\frac{mK\log \frac{1}{\delta}\log T}{\epsilon^2}\right)\,.  \qedhere\\
\end{align*}

\end{proof}

\section{Proof of Theorem 10}

\begin{proof}

Our lower bound is based on the following problem. There is $L$ items which all obey the Bernoulli distribution $P$. Each arm's weight is parameterized by:
\begin{align}
  \bar{w}(e) =
  \begin{cases}
    p & e \leq K \\
    p - \Delta & \text{otherwise}\,,
  \end{cases}
\end{align}
where $\Delta \in (0,p)$. Based on Lemma \ref{lem_4},

$$\mathbb{E}[R(A_t,w_t)] \geq \Delta(1-p)^{K-1} \sum_{e=K+1}^L \sum_{e^{\ast}=1}^K \mathbb{E}[\mathbbm{1}\{G_{e,e^{\ast},t}\}]\,.$$

\begin{equation}
    \begin{aligned}
        R(n)&\geq \Delta (1-p)^{K-1} \sum_{e=K+1}^L \mathbb{E}[\sum_{t=1}^n \sum_{e^{\ast}=1}^K \mathbbm{1}\{G_{e,e^{\ast},t}\}]\\
        &=\Delta(1-p)^{K-1} \sum_{e=K+1}^L \mathbb{E}[T_n(e)]\,.
    \end{aligned}
\end{equation}

By the conclusion from the work of Theorem 4 in \citet{basu2019differential}, we have that for any bandit
\begin{align*}
  \liminf_{n \to \infty} \frac{\mathbb{E}[T_n(e)]}{\log n} \geq \frac{1}{2 \min\{4,e^{2\epsilon}\}(e^\epsilon-1)^2d(p\Vert p-\Delta)}\,.
\end{align*}

Finally, using inequality $d(p\Vert p-\Delta) \leq \frac{\Delta^2}{p(1-p)}$, we get the final bound:

\[
  \liminf_{n \to \infty} \frac{R(n)}{\log n} \geq \frac{(L-K) p (1-p)^{K}}{2\Delta \min\{4,e^{2\epsilon}\} (e^{\epsilon}-1)^2}\,. \qedhere
\]

\end{proof}

\section{Proof of Theorem 11}
\begin{proof}
Our lower bound is based on the following problem. There is $L$ items which all obey the Bernoulli distribution $P$. Each arm's weight is parameterized by:
\begin{align}
  \bar{w}(e) =
  \begin{cases}
    p & e \leq K \\
    p - \Delta & \text{otherwise}\,,
  \end{cases}
\end{align}
where $\Delta \in (0,p)$. According to \citet{NEURIPS2018_a1d7311f}, the regret lower bound of bandit under DP has to suffer two terms: one term comes from the non-private bandit, one term comes from private bandit, as private cascading bandit is strictly harder than non-private cascading bandit (by reduction). The first term is at least $\Omega(\frac{(L-K)(1-p)^{K-1}\log n}{\Delta})$(\citet{kveton2015cascading}). 
Then we just have to prove cascading bandit under DP suffers regret at least $\Omega(\frac{(L-K)(1-p)^{K-1}\log n}{\epsilon})$.

\begin{align*}
        R(n)_{private} &\geq\sum_{t=1}^n \alpha \sum_{e=K+1}^L \sum_{e^{\ast}=1}^K \Delta_{e,e^{\ast}} \mathbb{E}_t[\mathbbm{1}\{G_{e,e^{\ast},t}\}]\\
        &\geq \Delta(1-p)^{K-1} \sum_{e=K+1}^L \mathbb{E}[T_n(e)]\,.
\end{align*}
Then according to Claim 14 from \citet{NEURIPS2018_a1d7311f}, for any sub-optimal arm, we get 
$$P\left[T_n(e)\geq \frac{\log n}{100\epsilon \Delta}\right] \geq \frac{1}{2}\,.$$
$$\mathbb{E}[T_n(e)]\geq\frac{1}{2}\times \frac{\log n}{100 \epsilon \Delta}=\frac{\log n}{200 \epsilon \Delta}\,.$$
So we get the conclusion:
\begin{align*}
    R(n)_{private} \geq\frac{ (1-p)^{K-1}(L-K)\log n}{200 \epsilon }=\Omega(\frac{(L-K)(1-p)^{K-1}\log n}{\epsilon})\,.
\end{align*}
Therefore, we get the final form of the result. \qedhere
\end{proof}

\section{Additional Experiments}

\begin{figure}[htbp] \label{fig}
\centering
\subfigure[LDP,$K=4$]{ 
\includegraphics[width=0.42\textwidth]{cascading_K_4_epsilon_02.pdf}
}
\subfigure[LDP, $K=8$]{  
\includegraphics[width=0.42\textwidth]{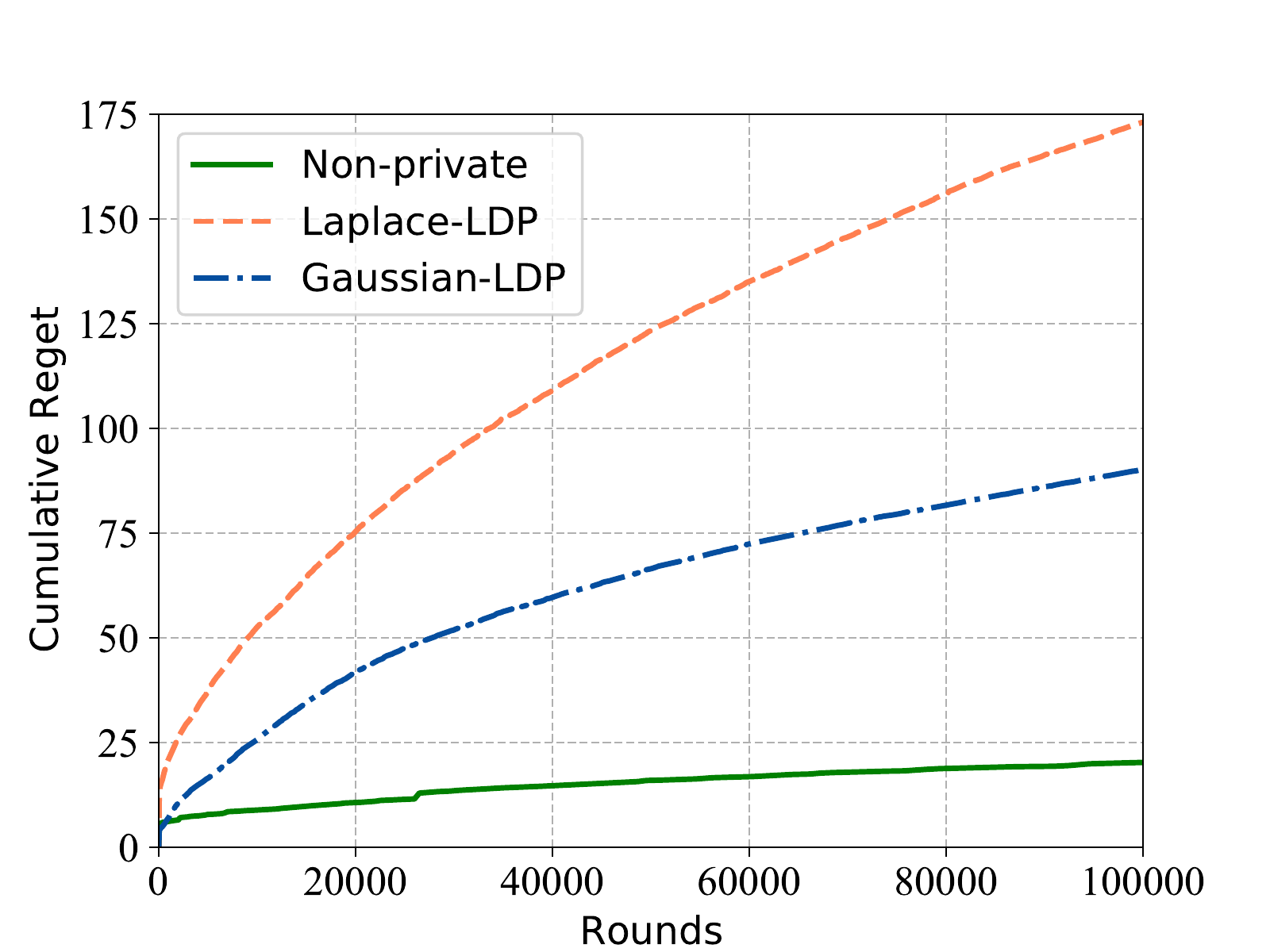}
}
\subfigure[LDP, $K=12$]{ 
\includegraphics[width=0.42\textwidth]{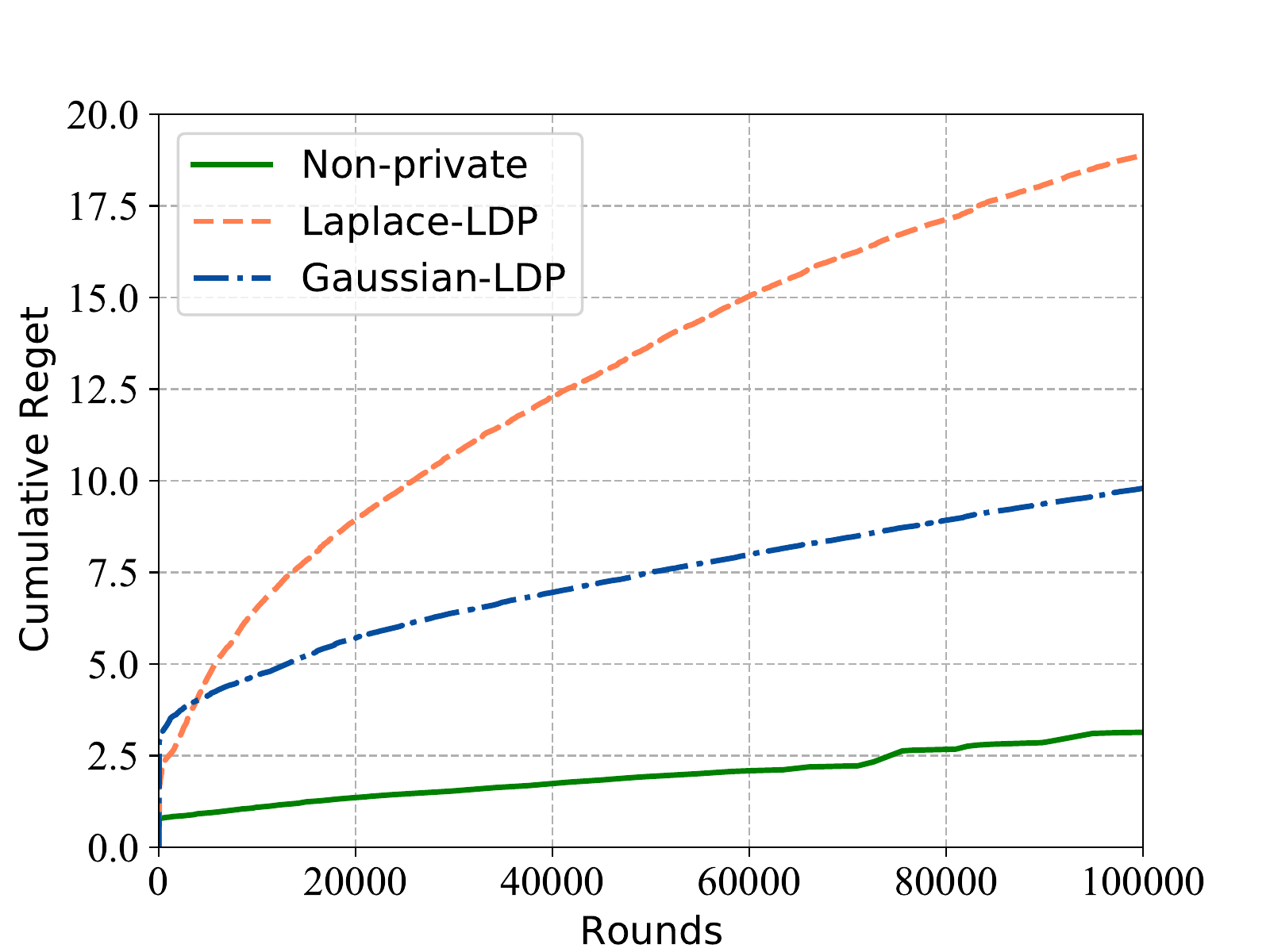}
}
\subfigure[LDP, $K=16$]{ 
\includegraphics[width=0.42\textwidth]{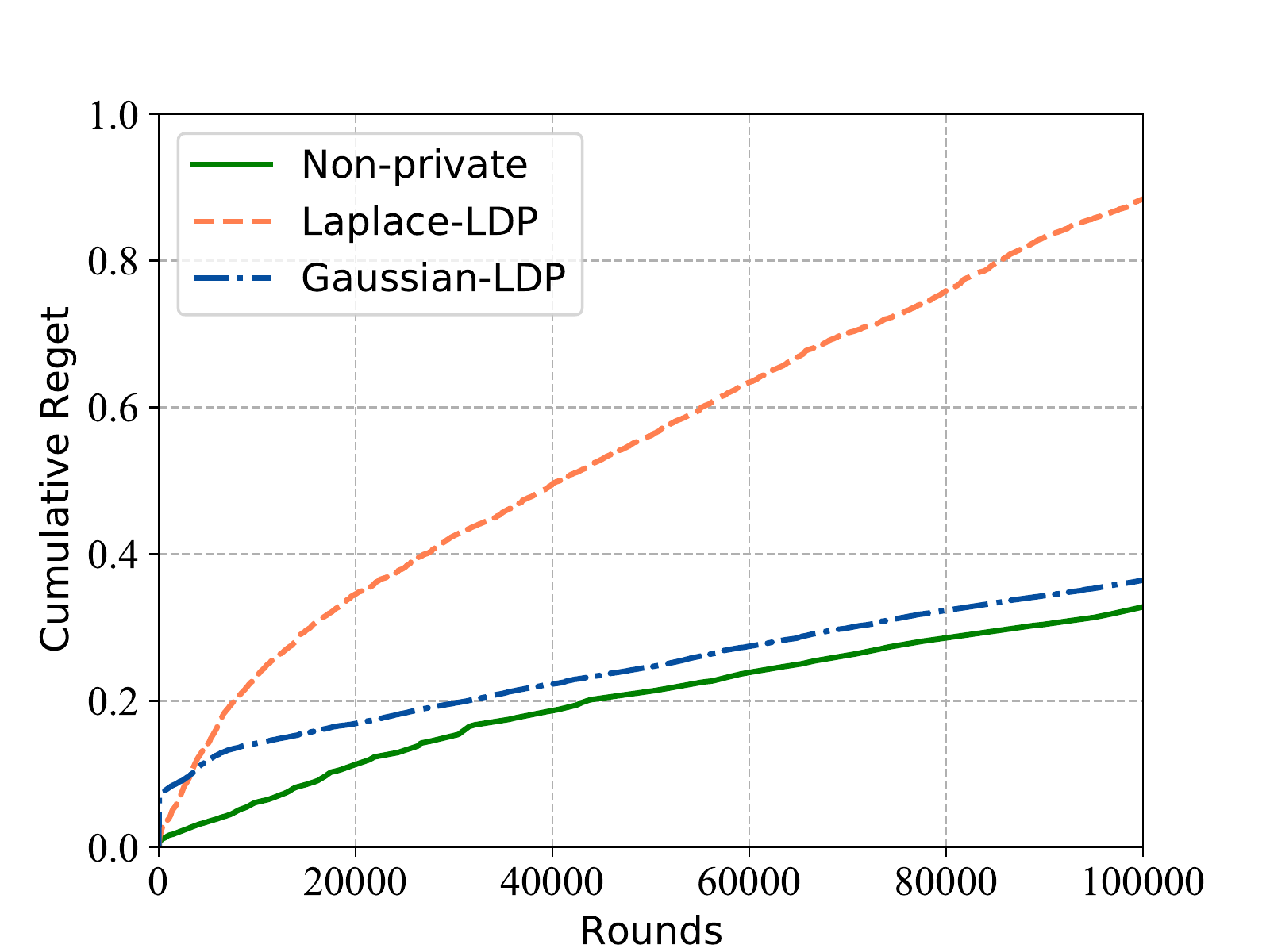}
}
\caption{Empirical results for private and non private algorithms under LDP when $\epsilon=0.2$. }
\end{figure}

In this section, we supplement some experiments with different number of arms under LDP. According to results, as $K$ increases, the gap between the Laplace mechanism and Gaussian mechanism becomes larger. This phenomenon reflects the influence of $K$ in the regret.

\end{document}